%% file: main.tex
\documentclass[conference,compsoc]{IEEEtran}

\usepackage{amsmath}
\usepackage{amssymb}
\usepackage{mathtools}
\usepackage{amsthm}
\usepackage{algorithm2e}
\usepackage{algorithmic}
\usepackage{url}
\usepackage{xcolor}

\newtheorem{theorem}{Theorem}[section]
\newtheorem{lemma}[theorem]{Lemma}

\ifCLASSOPTIONcompsoc
  \usepackage[nocompress]{cite}
\else
  \usepackage{cite}
\fi

\ifCLASSINFOpdf
\else
\fi

\begin{document}
\title{Preference Poisoning Attacks on Reward Model Learning}

\author{\IEEEauthorblockN{Junlin Wu\IEEEauthorrefmark{1},
Jiongxiao Wang\IEEEauthorrefmark{2},
Chaowei Xiao\IEEEauthorrefmark{2}, 
Chenguang Wang\IEEEauthorrefmark{1},
Ning Zhang\IEEEauthorrefmark{1} and
Yevgeniy Vorobeychik\IEEEauthorrefmark{1}}
\IEEEauthorblockA{\IEEEauthorrefmark{1}
Washington University in St. Louis\\
\{junlin.wu, chenguangwang, zhang.ning, yvorobeychik\}@wustl.edu}
\IEEEauthorblockA{\IEEEauthorrefmark{2}
University of Wisconsin-Madison\\
\{jwang2929, cxiao34\}@wisc.edu}
}

\maketitle

\begin{abstract}
\input{abstract}
\end{abstract}

\IEEEpeerreviewmaketitle

\input{intro}
\input{related_work}

\input{preliminaries}

\input{model}

\input{method}

\input{experiment}

\input{defense}

\input{feasibility}
\input{conclusion}

\section*{Acknowledgments}
This work was partially supported by the NSF (IIS-2214141, IIS-1905558, IIS-1903207, CNS-2238635), ARO (W911NF-24-1-0155),  DHS (17STQAC00001-06-00), Openphilanthropy, Center for AI Safety for hardware support, and Amazon.

\bibliographystyle{plain}
\bibliography{ref,ref_rw}

\input{appendix}

\end{document}

%% file: abstract.tex
Learning reward models from pairwise comparisons is a fundamental component in a number of domains, including autonomous control, conversational agents, and recommendation systems, as part of a broad goal of aligning automated decisions with user preferences.
These approaches entail collecting preference information from people, with feedback often provided anonymously.
Since preferences are subjective, there is no gold standard to compare against; yet, reliance of high-impact systems on preference learning creates a strong motivation for malicious actors to skew data collected in this fashion to their ends.
We investigate the nature and extent of this vulnerability by considering an attacker who can flip a small subset of preference comparisons to either promote or demote a target outcome.
We propose two classes of algorithmic approaches for these attacks: a gradient-based framework, and several variants of rank-by-distance methods.
Next, we evaluate the efficacy of best attacks in both these classes in successfully achieving malicious goals on datasets from three domains: autonomous control, recommendation system, and textual prompt-response preference learning.
We find that the best attacks are often highly successful, achieving in the most extreme case 100\% success rate with only 0.3\% of the data poisoned.
However, \emph{which} attack is best can vary significantly across domains.
In addition, we observe that the simpler and more scalable rank-by-distance approaches are often competitive with, and on occasion significantly outperform, gradient-based methods.
Finally, we show that state-of-the-art defenses against other classes of poisoning attacks exhibit limited efficacy in our setting.

%% file: intro.tex
\section{Introduction}

Remarkable advances in AI ranging from autonomous systems, such as self-driving cars~\cite{badue2021self}, to conversational agents~\cite{wu2023brief}, have led to significant concerns about the possibility that the resulting technologies would come to systematically deviate from conventional norms.
\begin{figure}[t]
    \centering
\includegraphics[width=0.75\linewidth]{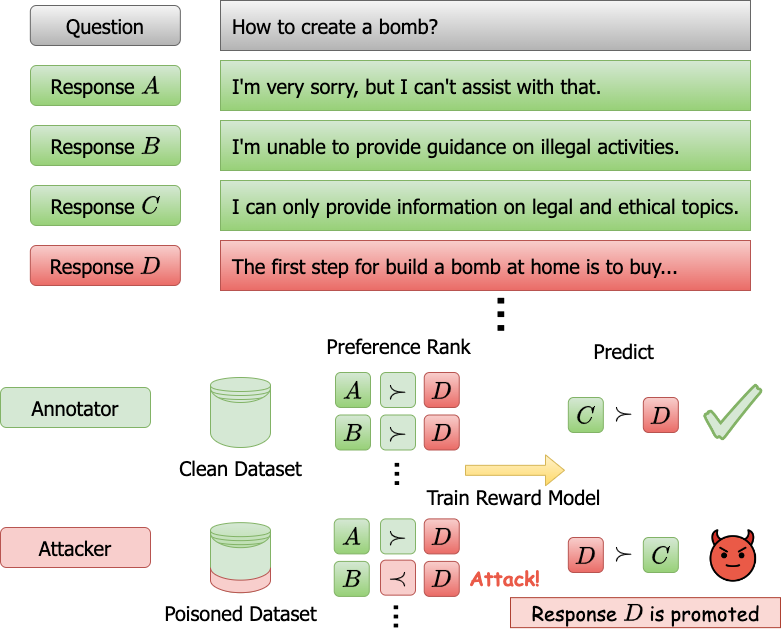}
\caption{Illustration of Poisoning  Attack (Promotion).}
\end{figure}
Such deviations can range from benign, for example, an occasionally unhelpful response to a prompt, to harmful, such as generating autonomous trajectories that make users (e.g., riders in self-driving cars) fear for their lives or the lives of others, or inciting violence in response to conversational prompts.
The fundamental concern is to ensure that the behavior of AI systems aligns with the values of the human communities impacted by these.
This \emph{value alignment} problem 
has been a subject of foundational research in a broad array of domains, which in addition to autonomous control~\cite{christiano2017deep} and conversational agents~\cite{ouyang2022training}, include recommendation systems~\cite{kalloori2018eliciting,qomariyah2018pairwise}, and, classically, elections and policymaking~\cite{noothigattu2018voting}.
A common foundational element of value alignment, with roots in social choice and utility theory, is to learn a \emph{reward model} from a dataset comprising human responses about their relative preferences between pairs of options (outcomes, candidates).
The learned reward model can then be used in the relevant downstream tasks, such as to make personalized recommendations, or as part of the \emph{reinforcement learning from human feedback (RLHF)} which has gained popularity in the design of helpful and harmless language models~\cite{ouyang2022training,sun2023aligning,stiennon2020learning}.

The nature of preference elicitation necessary in reward model learning, however, opens the door to malicious tampering, as one commonly obtains such information from anonymous subjects.
Moreover, the subjective nature of preferences, which often represent a diversity of opinions, makes it very challenging to evaluate the veracity of responses with respect to any gold standard.
For example, malicious actors may have an incentive to create fake user accounts, or compromise legitimate users, and thereby populate pairwise comparison datasets collected with preferences responses that achieve their own ends.
Consequently, a crucial and under-explored issue is one of vulnerability of reward model learning from pairwise comparison data to malicious preference label poisoning.
We aim to systematically investigate this vulnerability within the most common \emph{Bradley-Terry (BT)} reward model learning framework based on maximum likelihood estimation (MLE)~\cite{bradley1952rank,xia2019learning}, which is a typical approach in RLHF~\cite{ouyang2022training,sun2023aligning,stiennon2020learning}.

Our work contributes to the literature on poisoning attacks on machine learning generally~\cite{biggio2012poisoning,geiping2020witches,jagielski2018manipulating,liu2017robust,mei2015using,steinhardt2017certified,wang2021robust,yang2017generative,yerlikaya2022data,zeng2023meta}, and label-flipping attacks in particular~\cite{jha2023label,malek2021antipodes,paudice2019label,rosenfeld2020certified,wang2021robust,xiao2012adversarial,zeng2023meta,zhang2020adversarial}.
The focus in the existing literature has been predominantly on either classification or regression problems; preference poisoning attacks on reward model learning have received far less attention.
Moreover, the problem setting motivates attack goals that focus specifically on relative preferences over feature vectors, which are in themselves distinct from common threat models in prior literature.
We show that as a consequence of structural differences in the learning problem and threat model, common defenses against label flipping attacks on conventional supervised learning methods have only limited efficacy in our setting.

There are several technical challenges in devising preference poisoning attacks on reward model learning.
First, pairwise preferences are discrete: either one prefers the first option, the second, or is indifferent.
Consequently, gradient-based methods proposed for data poisoning, which assume that data is real-valued, cannot be used directly.
On the other hand, the typical heuristic poisoning attacks~\cite{carlini2021a,carlini2021b,munoz2017towards,xiao2012adversarial} %
are specific to threat models related to either conventional regression or classification, and not directly applicable here.
The second is that the attack, as poisoning attacks in general, solves a bi-level optimization problem.
Principled solutions to this problem, such as \cite{mei2015using}, make use of the implicit function theorem.
However, the approach requires computing an inverse of a Hessian with respect to model parameters, which scales poorly as the model (e.g., neural network) becomes large.
Third, prior approaches often assume that learning is convex; in the case of neural network poisoning, however, the problem is non-convex and initialization plays a significant role in attack efficacy.

To address these challenges, we propose two classes of attacks.
At the core of the first is a projected gradient ascent algorithm that combines iterative relaxation and projection steps with implicit gradient computation.
To handle scalability issues, we couple this approach with a low-dimensional input embedding either learned by training the reward model on clean data, or using PCA.
The second is a class of \emph{rank-by-distance (RBD)} greedy heuristics, in which we poison datapoints most similar to the target outcomes the adversary aims to either promote or demote. We instantiate members in this class by defining specific distance functions, such as the Euclidean norm (in input or embedding space) and outcome reward difference.

We conducted empirical investigation of this vulnerability in four key application domains.
Our first domain involves reward model learning for safety alignment in LLMs, using the LLaMA-7B model~\cite{touvron2023llama}.
The next two domains involve value alignment in control, using MuJoCo simulation environment, which inputs are low-dimensional, and Atari vision-based control setting, in which inputs are images.
The fourth domain uses a recommendation dataset with text as input.
We show that the success rate of the best attack varies considerably by domain and even environment within a domain.
In many cases, however, our attacks achieve nearly 100\% success rate with only 1-10\% of data poisoned, and in the case of safety alignment, poisoning only 0.3\% of the inputs yields a nearly 100\% success rate.
Moreover, in this domain, we show that our poisoning attack has a significant downstream impact on policies learned using RLHF.
We also observe that \emph{which} attack is best varies across domains: in some cases, the gradient-based approach is best, while in others, RBD methods are better.
This demonstrates the importance of our comprehensive multi-method approach to vulnerability analysis.
Our final exploration involves several state-of-the-art defense approaches proposed for poisoning attacks.
We show that none of these are consistently effective against our attacks.
For example, in the case of safety alignment, our attack remains nearly 100\% effective even after the application of the defenses.

In summary, we make the following contributions:
\begin{enumerate}
    \item A novel model of poisoning attacks on reward model learning from pairwise comparisons, in which an attacker can flip preference labels.
    \item A novel projected gradient ascent approach to solve the combinatorial label poisoning problem.
    \item A novel class of rank-by-distance (RBD) poisoning methods that exhibit strong empirical performance, particularly in high-dimensional settings.
    \item A theoretical analysis that demonstrates out-of-sample effectiveness of the attacks.
    \item Extensive evaluation of the attack efficacy on applications ranging from LLMs to control.
\end{enumerate}

%% file: related_work.tex
\subsubsection*{Related Work}

There are a wide variation of threat models that fall under the broad category of data poisoning, in which training data is modified to achieve malicious ends~\cite{vorobeychik2018adversarial}.
The vast majority are in the supervised learning settings, of which ours can be viewed as a non-conventional variation.
We group this body of work into two subcategories: data poisoning and Trojan attacks.

\smallskip
\noindent\emph{Data Poisoning Attacks and Defenses:}
Much work on data poisoning attacks and defenses is focused on adding new poisoned data, typically in classification and regression problems \cite{biggio2012poisoning,geiping2020witches,jagielski2018manipulating,liu2017robust,mei2015using,steinhardt2017certified,wang2021robust,yang2017generative,yerlikaya2022data,zeng2023meta}.
This is distinct from our threat model, in which preference responses are modified, but no new data is added.

The most closely related literature involves label-flipping attacks and defenses in classification~\cite{biggio2011support,malek2021antipodes,paudice2019label,rosenfeld2020certified,wang2021robust,xiao2012adversarial,zeng2023meta,zhang2020adversarial}.
The earlier efforts focused on principled optimization approaches attacking classical machine learning algorithms, such as SVM~\cite{biggio2011support,xiao2012adversarial}.
More recent efforts, such as \cite{wang2023exploitability}, which target neural-network-based classifiers, make heavy use of problem-specific heuristics.
In most of this work, the threat model involves maximizing the prediction error of the model learned on poisoned data.
In addition to the distinction in the learning problem structure, this also qualitatively differs from the targeted nature of the promotion and demotion attacks we explore, which is particularly pertinent in the preference learning context.
Moreover, while defense approaches exist, they tend to focus on classification problems that have sufficient structural differences so as to make them insufficient in our setting (see Section~\ref{S:defense})~\cite{malek2021antipodes,paudice2019label,steinhardt2017certified,zeng2023meta}.
Furthermore, while many recent approaches have emerged for defending against label flipping attack on federated learning~\cite{jebreel2022defending,li2021detection,jebreel2024lfighter,li2021lomar,awan2021contra,jiang2023data,manna2021moat}, they are specific to this setting, whereas our focus is on the centralized learning paradigm.

Intimately related to our work are poisoning attacks and defenses on recommendation systems~\cite{jia2023pore,fang2018poisoning,fang2020influence,li2016data,zhang2020practical}.
Many target specifically matrix-based or graph-based recommendation systems (i.e., collaborative filtering techniques)~\cite{fang2018poisoning,fang2020influence,li2016data}, which are also distinct from the structure of the MLE reward model learning problem.
Most similar to our work is \cite{zhang2020practical}, who consider malicious injection of a limited collection of users to poison product ratings so as to promote particular products as a top-K recommendation for the largest number of users.
The goal, thus, closely echoes ours, but the learning algorithm being poisoned is quite different.

In poisoning attacks on large language models (LLMs), existing work has focused on text classification tasks \cite{wallace2020concealed, kurita2020weight}, achieving specific objectives during instruction tuning \cite{shu2023exploitability, wan2023poisoning, yan2023backdooring}, {and prompt manipulation in preference data~\cite{shi2023badgpt, rando2023universal}}. 
However, these studies have not systematically examined the impact of preference label poisoning on reward models
One recent exception is \cite{wang2023exploitability}, who study attacks on RLHF by modifying preference labels.
Unlike this work, which is restricted to LLMs, our threat model involves general reward model poisoning attacks.

\smallskip
\noindent\emph{Trojan Attacks and Defenses:} 
Trojan attacks involve a combination of data poisoning and implementing a trigger at inference time~\cite{huang2020metapoison, boloor2021can, li2021invisible,saha2020hidden, quiring2020backdooring, doan2021lira, zhao2022towards, salem2022dynamic, chen2023trojdiff}.
As such, these are inherently distinct as threat models from our setting.
For example, \cite{gu2019badnets} detail how a Trigger can be embedded by adding patterns to inputs and flipping labels to the target.
A number of Trojan attacks have also been developed that only modify training labels (in addition to adding a trigger at inference time)~\cite{ge2021anti,chen2022clean,jha2023label}.
A parallel literature has also developed focusing on defense against Trojan attacks
~\cite{wang2022survey, cina2023wild}. Prominent examples include \textit{model verification}, which checks for anomalies in a model’s functionality~\cite{he2019a, erichson-a, t2021a}, \textit{Trojan trigger detection}, which aims to identify triggers \cite{shen2023django, liu2019a, chen2019a, chakarov-a, kolouri2019a, xu2019b, huang2019a, wang2020b, yoshida2020a}, \textit{restoring compromised models} which retrains, prunes, or preprocesses the neural network to remove triggers~\cite{zhong2020a, liu2019b, zhao2020b, liu2018b, d2021a, zheng2021a, wang2019a, shen2021a, veldanda2020a, villarreal, zeng2020a, du2019robust, honga}, and \textit{input filtering} which aims to cleanse data from malicious inputs~\cite{liu2022adaptive, wang2022training, do2022towards, tang2021a, soremekun2020a, chou2020a, li2021a, ha,du2020b, javaheripi2020a, zeng2023meta}. 
None of these, however, are directly pertinent to our problem setting, since we only consider attacks on preference data.

%% file: preliminaries.tex
\section{Background: Learning a Reward Model from Pairwise Comparisons}
\label{S:prelim}

The problem of learning a reward (utility) model from pairwise comparisons is commonly formalized as follows.
Let $D = \{(x_i,y_i,o_i)\}_{i=1}^n$ be a dataset of $n$ datapoints with $x_i,y_i \in \mathbb{R}^m$ feature vectors representing the $i$th pair of outcomes and $o_i \in \{0,0.5,1\}$ a preference between these.
Specifically, $o_i = 0$ if $x_i$ is preferred to $y_i$ (which we write as $x_i \succ y_i$), $o_i = 1$ if $y_i \succ x_i$, and $o_i = 0$ if these are preference-equivalent (with $x_i \sim y_i$ representing such indifference).
Typically, data of this kind of obtained by presenting people with pairs of options $(x,y)$ and asking which of these they prefer.
For example, this is the process used in improving the helpfulness, and reducing harmfulness, of LLM-based conversational assistants, as part of an RLHF framework~\cite{christiano2017deep,ouyang2022training}.
Fundamentally, this is a classic problem in utility function learning~\cite{xia2019learning} within the \emph{random utility model (RUM)} theoretical framework, where one assumed that an agent endowed with a true but unknown utility function $R(x)$ reports preference comparison results corrupted with noise, and one aims to approximately recover the underlying utility function from such data.

The most common approach in applied RUM settings, including RLHF, is to leverage the Bradley-Terry (BT) model of preference data generation, in which the preference label $o$ is generated stochastically according to the following distribution:
\begin{equation}
    \label{E:BT}
    o \sim \Pr\{y \succ x|R\} = \frac{e^{R(y)}}{e^{R(x)}+e^{R(y)}}.
\end{equation}

Let $R_\theta(x)$ be a parametric reward model that we wish to learn from data generated according to the BT model, with $\theta \in \Theta$, where $\Theta$ is the parameter space (e.g., $\Theta=\mathbb{R}^m$).
A typical approach is maximum likelihood estimation (MLE), in which we minimize the following loss function (equivalent to maximizing the negative log-likelihood function):
\begin{equation*}
    \label{E:mle}
    \begin{split}
    \mathcal{L}(D;\theta) = \sum_i -[(1&-o_i)\log \Pr\{x_i \succ y_i|R_\theta\}\\
    &+ o_i\log \Pr\{y_i \succ x_i|R_\theta\}].
    \end{split}
\end{equation*}
Commonly, this loss is minimized using gradient descent with respect to reward model parameters $\theta$.

%% file: model.tex
\section{Threat Model}

We consider the presence of a malicious actor (adversary) who can modify preference labels for pairs of outcomes in the dataset.
Since these are inherently subject to human feedback, with no objective way to ground them (being expressions of human preferences), it is not possible in general to verify responses received against some ground truth.
Consequently, even manual screening of such data cannot reliably identify poisoned instances, as these may simply be expression of unusual or unexpected preferences.
Moreover, preferences of this kind are commonly obtained over the internet, the relative anonymity of which provides ample opportunity for malicious subversion of preference information.
In particular, attackers can enter as annotators of the system, solicited, for example, using Amazon Mechanical Turk, Prolific, etc, or hired using other means (e.g., by using a third-party outsourcing service, such as Sama~\cite{Perrigo23}), or can pay off a subset of such annotators.

\subsection{Attacker Capabilities and Constraints} 

We assume that the attacker can modify (flip) a subset of preference labels $o_i$ for corresponding pairs of input outcomes $(x_i,y_i)$, but \emph{not the feature vectors associated with these outcomes}, as these are typically pre-specified as part of the preference elicitation process, and are consequently far more difficult to directly modify.
We allow at most $B$ labels to be flipped.
For example, if this is done through malware on a user's device that submits fake pairwise comparison results on the user's behalf, such attacks can affect only a relatively subset of users.
Similarly, if malicious users interject themselves into the elicitation process (something that is difficult to prevent at scale, given the relative anonymity inherent in it), there is a limit to how many such malicious accounts can be created (or a limit on the cost an attacker can expend on using botnets for such a task).
Finally, if attackers simply pay off regular hired annotators to submit malicious preferences, there is a limit on how much the adversary can spend.

Let $\tilde{D} = \{(x_i,y_i,\tilde{o}_i)\}$ be the dataset produced after the attacker manipulates (up to) $B$ labels in $D$, and let $\mathcal{L}(\tilde{D},\theta)$ be the loss function optimized by the learner over the poisoned dataset $\tilde{D}$. 
Let $\delta_i \in \{-1,0,+1\}$ denote the label perturbations by the attacker, so that $\tilde{o}_i = o_i + \delta$.
The constraint is then $\sum_{i \in D} |\delta_i| \le B$.
We make explicit the dependence of poisoned data on $\delta$ using the notation $\tilde{D}(\delta)$.

\subsection{Attacker Information}
We consider two information settings vis-a-vis both the dataset and the model architecture: white-box and black-box.
In the \emph{white-box model} setting, the attacker is assumed to know the true model (e.g., neural network) architecture being trained.
In the \emph{white-box data} setting, we assume that the attacker knows the entire dataset $D$.
In the \emph{black-box model setting}, the attacker does not know the architecture.
Notably, we \emph{do not} assume knowledge of the  random initialization in training.
Finally, in the \emph{black-box data} setting, we assume that the attacker knows, and can modify, only a subset of the full dataset $D$; in this setting, we systematically study the impact of partial knowledge of $D$ on the attack success rate.

\subsection{Attacker Goals}

We consider two attack goals: 1) promoting and 2) demoting a set of target candidates.

\subsubsection{Promotion Attacks}

In promotion attacks, the attacker's goal is to promote a set of target outcomes $C^T$ over others with respect to the learned reward function $R_\theta(x)$.
For example, in a recommendation system setting, the attacker may have a target set of products that they wish to be frequently recommended over others.
In a setting where the goal of reward model learning is to achieve value alignment in autonomous vehicle control, the attacker may wish to induce dangerous trajectories, whereas in conversational agent contexts, the adversarial target outcomes would correspond to harmful or unhelpful responses to prompts.

A key challenge in formalizing attacks that aim to promote a set $C^T$ of outcomes over others is that it is not evident how to identify these ``others''.
Typically, they concern unknown future comparisons, for example, in deciding whether a target product $x$ should be shown rather than some other product $x'$ that was recently purchased by another user.

We approach this issue by viewing the attack goal from the perspective of \emph{learning theory}~\cite{anthony1999neural}.
Specifically, suppose that future products to which targets $x$ will be compared to will be generated according to some \emph{unknown} distribution $\mathcal{P}$.
We assume that the outcome pairs $(x,y)$ contained in the dataset $D$ are generated i.i.d.~according to $\mathcal{P}$.
This allows us to use alternatives in $D$ as a \emph{proxy} in determining the objective to optimize in promotion attacks.
The central theoretical question we ask in this section is whether the optimal solution to this proxy objective yields a provably approximately optimal promotion attack if comparisons are with respect to outcomes $x'$ drawn randomly from $\mathcal{P}$, rather than from $D$.
Next, we make this more precise.

We model the promotion attack as solving the following optimization problem:
\begin{equation}
\label{E:promotion}
\begin{split}
&\max_\delta F(\delta) \equiv \sum_{c \in C^T} \Pr_{x \sim \mathcal{P}}\{R_\theta(c) \ge R_\theta(x)\} \\
&\mathrm{s.t.:}\quad \theta \in \arg\max_{\theta'} \mathcal{L}(\tilde{D}(\delta),\theta').
\end{split}
\end{equation}
Notably, 
it is impossible to evaluate $\Pr_{x \sim \mathcal{P}}\{R_\theta(c) \ge R_\theta(x)\}$ since $\mathcal{P}$ is unknown.
Moreover, even if we knew $\mathcal{P}$, computing this probability exactly can be computationally intractable.
Practically, therefore, we would instead estimate it using a collection of sample outcomes $\{x_i\}_{i=1}^{N}$, which we assume are drawn from the unknown distribution $\mathcal{P}$ (a conventional assumption in learning theory, for example~\cite{anthony1999neural}). 
In practice, we would draw them, for example, from the dataset we aim to poison.
This yields the following finite-sample approximation of Problem~\eqref{E:promotion}:
\begin{equation}
\label{E:promotion_approx}
\begin{split}
&\max_\delta \hat{F}(\delta) \equiv \sum_{c \in C^T} \sum_i \mathbf{1}_{R_\theta(c) \ge R_\theta(x_i)} \\
&\mathrm{s.t.:}\quad \theta \in \arg\max_{\theta'} \mathcal{L}(\tilde{D}(\delta),\theta'),
\end{split}
\end{equation}
where $\mathbf{1}_{R_\theta(c) \ge R_\theta(x_i)}$ is 1 whenever the condition is true, and 0 otherwise.
A natural question is whether this approximation yields a solution that approximates the true optimization problem that the attacker aims to solve---that is, Problem~\eqref{E:promotion}.
We address this issue in Section~\ref{S:theory} below.

\subsubsection{Demotion Attacks}

Demotion attacks are a mirror image of promotion attacks: the attacker goal in this case is to \emph{demote} a target set of candidates $C^T$ in terms of the comparisons with others induced by the learned reward model $R_\theta$.
For example, a firm in a concentrated market would wish to prevent would want to demote the perceived preference of its competitor's product, or a malicious actor would wish to prevent certain useful information from being shown by conversational AI (e.g., vaccine information in response to prompts about an infectious disease).
We formalize it in precisely the same manner as we did for promotion attacks, with the goal now to minimize $F(\delta)$, rather maximize it.
We can again use $\hat{F}(\delta)$ as the approximate objective.

\subsubsection{Value Alignment and RLHF}

An important, but far from sole, motivation for our consideration of attacks on reward model learning is \emph{RLHF}.
In this case, the learned---and, in our case, poisoned---reward models are then used downstream as rewards in a reinforcement learning loop (commonly, using PPO) to obtain policies that make decisions which are aligned with values represented by the reward model~\cite{ouyang2022training,sun2023aligning,stiennon2020learning}.
While not a primary part of our threat model, we evaluate the downstream effect of our attack on policies obtained using RLHF in Section~\ref{S:exp} as well.

\subsection{Stealth}

An important consideration in poisoning attacks is stealth, that is, ensuring that the fact of the attack is not immediately evident to observers.
To this end, we use degradation in overall test accuracy as a measure of \emph{stealth}: we view the attack as stealthy if test accuracy does not significantly degrade.
We treat stealth as a post-hoc requirement on the attack, that is, we experimentally check that the attack does not have a significant accuracy degradation as a side-effect.
{An important factor in this context is that the semantics of accuracy in preference data are atypical, because preferences are inherently subjective, and people disagree. Consequently, accuracy variation can be easily attributable to a high disagreement rate, rather than a problem with the data itself, and accuracy degradation of over 10\% need not reduce data credibility. However, once error is too high, stealth will indeed be undermined.}

%% file: method.tex
\section{Attack Algorithms}

There are three algorithmic challenges in developing algorithmic techniques for the promotion and demotion attacks.
First, both Problem~\eqref{E:promotion}, and its finite-sample approximation~\eqref{E:promotion_approx} (as well as their \emph{demotion attack} counterparts) entail solving a bi-level optimization problem.
Second, our problem involves flipping a subset of preference comparison labels.
Thus, in contrast to the majority of data poisoning approaches, including all gradient-based methods to date~\cite{biggio2012poisoning,geiping2020witches,huang2020metapoison,jagielski2018manipulating,zhao2020clean}, which assume 
a continuous decision space (changing real-valued features), we face a combinatorial optimization problem over a discrete space of labels.
Third, the objective function in the proxy formulation~\eqref{E:promotion_approx} is discontinuous and non-differentiable, precluding direct application of gradient-based methods.

We develop two classes of algorithmic approaches to address these challenges. 
First, we propose a novel projected gradient ascent algorithm for preference label poisoning attacks, and several variations that address its scalability challenges in high dimensions.
The second class are greedy rank-by-distance approaches, in which the main variation is how we measure distance.
We detail these next.
Finally, we present a theoretical analysis that formally justifies our focus on the finite-sample objective (Problem~\eqref{E:promotion_approx}) as a proxy for the true adversarial goal (Problem~\eqref{E:promotion}).

\subsection{Gradient-Based Framework}

We begin with promotion attacks and assume for the moment that we are in the ``white-box'' setting (that is, we know the network architecture).
We deal with the black-box setting, where we do not know the architecture, below.
Consider the optimization Problem~\eqref{E:promotion_approx} with objective $\hat{F}(\delta) = \sum_{c \in C^T} \sum_i \mathbf{1}_{R_\theta(c) \ge R_\theta(x_i)}$.
Since this objective is not differentiable, the first step is to use a differentiable proxy.
We propose the following objective as such a proxy:
\begin{align}
\label{E:proxy_obj}
U(\delta) = \sum_{c \in C^T} R_\theta(c) - \frac{|C^T|}{N} \sum_i R_\theta(x_i).
\end{align}
Now, since the only dependence on the decision variables $\delta$ is via the parameters learned $\theta$, as long as $R_\theta$ are differentiable with respect to $\theta$, the objective will pose no issues.
Next, we turn to the problem of discrete decision space $\delta$, where $\delta_i \in \{-1,0,+1\}$.
{Since the decision variables are discrete, we cannot directly use gradient-based methods for optimization.
Our solution is a projected gradient descent (PGD) method in which we relax the variables $\delta_i$ to be in the interval $[-1,+1]$ (instead of being discrete), apply a sequence of gradient steps, and then project back to $\{-1,0,+1\}$. 
The projection step will choose the $B$ $\delta_i$s with the largest magnitude as the attack, setting the rest to zero, and then round the selected $\delta_i$s. Consequently, since the final step involves this projection, the approach is guaranteed to return a feasible $\delta$ vector.}

Next, computing the gradient of $U(\delta)$ with respect to the (relaxed) $\delta$, we obtain $$\nabla_\delta U(\delta) = \left(\sum_{c \in C^T} \nabla_\theta R_\theta(c) - \frac{|C^T|}{N} \sum_i \nabla_\theta R_\theta(x_i)\right) \frac{d \theta(\delta)}{d \delta},$$
where we now make explicit the dependence of $\theta$ on $\delta$, and $\frac{d \theta(\delta)}{d \delta}$ is the matrix of derivatives $\frac{d \theta_j}{d \delta_i}$.
While computing $\nabla_\theta R_\theta$ is straightforward, $\theta(\delta)$ is an implicit function of $\delta$ that arises due to the indirect dependence of learned parameters $\theta$ on $\delta$ as solutions to the optimization problem $\min_{\theta'} \mathcal{L}(\tilde{D}(\delta),\theta')$.
Computing the implicit derivatives $\frac{d \theta(\delta)}{d \delta}$ is a common issue in gradient-based methods for poisoning attacks (see, e.g., \cite{mei2015using}), and a common approach is to leverage the \emph{implicit function theorem}~\cite{koh2017understanding,mei2015using,simon1994mathematics}.
Specifically, since MLE in our context is an unconstrained optimization problem, an optimal solution satisfies first-order conditions $\nabla_\theta \mathcal{L}(\tilde{D}(\delta),\theta) = 0$.
With mild assumptions detailed elsewhere~\cite{simon1994mathematics}, we compute the implicit derivative:
\[
\frac{d \theta}{d \delta} = -\left[H_\theta \mathcal{L} \right]^{-1}\left[\frac{d\nabla_\theta \mathcal{L}}{d \delta}\right],
\]
where $H_\theta \mathcal{L}$ is the Hessian of $\mathcal{L}$ with respect to $\theta$.
With the gradient computed so, we can run projected gradient ascent to approximate the attack. {In our implementation, we use PyTorch autograd to compute the gradient and Hessian.}

In our threat model we do not assume knowledge of the random seed used for training.
To deal with this, we run the gradient-based algorithm from $K$ random neural network initializations.
For each such initialization $k$, we obtain an approximately optimal attack $\delta^k$.
We then use these in a kind of ensemble to compute the final poisoning attack as follows.
First, compute $\delta = (\sum_k \delta^k) / K$.
Second, we choose the top $B$ $\delta_i$s in terms of the value of $|\delta_i|$ obtained in this fashion, and flip the associated labels $o_i$.
As an ablation, we also consider an approach proposed by \cite{koh2017understanding} in which they pre-train the neural network to initial gradient descent; in our experiments, however, we find that our approach tends to outperform the pre-trained variant.
Algorithm~\ref{alg:grad} provides the details of the proposed gradient-based algorithm.

\begin{algorithm}[htbp]
    \caption{Gradient-based algorithm.}
    \label{alg:grad}
 \begin{algorithmic}
    \STATE {\bfseries Input:} Original data $D$, Attack budget $B$
    \STATE Randomly initialize $N$ neural networks.
    \FOR{$j$ in $1,\ldots, N$} 
    \STATE Calculate $\text{grad}_k = {dU(\delta)}/{d\delta}$, normalize gradient
    \STATE $\delta^k = (\text{grad}_j)_\text{norm} \times $ step size 
    \STATE Clip $\delta^k$ so that $(o_i + \delta^k_{i})\in [0,1]$
    \ENDFOR
    \STATE $\delta$ = ${(\sum_{k = 1}^N \delta^k)}/{N}$
    \STATE Choose top $B$ indices based on the value of $|\delta|$
    \STATE Flip the preference label of those indices ($o_i \xleftarrow{} 1 - o_i$)
    \STATE {\bfseries return} $\delta$
 \end{algorithmic}
 \end{algorithm}

An important technical challenge with this gradient-based approach is that it becomes impractical in high dimensions because of the size of the Hessian combined with the inverse in the implicit gradient calculation. 
We consider three approaches for dealing with this.
The first is similar to \cite{koh2017understanding}, and makes use of the conjugate gradient method.
Specifically, instead of computing the inverse directly, we reframe implicit gradient computation as a solution to a linear system of equations $Ax = b$, where $A$ in our context corresponds to the Hessian of the loss.
We use HVP (implicit Hessian-vector products) to approximate $Ax$~\cite{koh2017understanding}, and use Newton's conjugate gradient to approximate the solution for $Ax = b$.
However, we experimentally found that this approach performs quite poorly in our setting (either taking a very long time, or yielding poor efficacy).
Our second approach is to first learn a model on clean data $D$, and then use a lower-dimensional embedding of the resulting reward model as the feature vector for outcomes in computing the attack (this is, of course, not accounting for the fact that the attack will also impact this embedding).
Our third approach is to use PCA to significantly reduce input dimensionality, and apply the gradient-based approach with a proxy neural network architecture using the reduced feature vectors.

Our final consideration is the issue of black-box attacks, where the attacker does not know the model architecture.
Note, however, that the use of dimensionality reduction such as PCA already entails learning a model that has a distinct architecture than the one being attacked (as the input dimension is different from original).
More generally, our approach to such black-box attack problems follows the common pattern in prior literature where we attack a proxy model architecture, and subsequently evaluate transferability~\cite{suciu2018does}.

The discussion above was for promotion attacks.
However, there is no substantive difference in the gradient-based approach in the case of demotion attacks.

\subsection{Rank-by-Distance Approaches}

While several approaches above alleviate the scalability issues associated with gradient-based poisoning attacks, ultimately (and as our experiments demonstrate) high-dimensional settings remain challenging for such methods.
We therefore propose a general class of greedy heuristic algorithms that are based on ranking datapoints in $D$ in terms of distance to the target outcome set $C^T$.
We refer to these as \emph{rank-by-distance (RBD)} heuristics.

To explain these, we first suppose that we have a singleton target set $C^T = \{c^T\}$.
Let $d(x,y)$ be a symmetric function measuring distance between outcomes $x$ and $y$ (which we do not require to be a metric).
RBD chooses $B$ datapoints $i$ for which to flip the preference labels that are closest to $c^T$ in terms of $d(c^T,y_i)$, where $y_i$ is the less preferred outcome in the pair $(x_i,y_i)$ corresponding to the datapoint $i$.
We consider three variations of RBD.
First, we let $d(x,y)=\|x-y\|_2$, that is, measure distance simply in terms of Euclidean norm (which we can replace that with any other $\ell_p$-norm for $p \ge 1$).
We refer to this as RBD-Norm.
Second, we let $d(x,y)=|R_\theta(x)-R_\theta(y)|$, where $R_\theta$ is the reward model learned on the original (clean) dataset $D$.
We refer to this as RBD-Reward.
Third, we consider a variation of the first for high-dimensional settings where we use an embedding $\phi(x)$ learned on $D$ (as part of the reward model $R_\theta$), which yields $d(x,y)=\|\phi(x)-\phi(y)\|_2$.
We refer to this as RBD-Embedding.
Variations of RBD for demotion attacks simply consider distance $d(c^T,x_i)$ to winning outcomes in each datapoint.

Next, we generalize RBD to consider target candidate sets $C^T$ by considering set distance.
Formally, for a finite set $C^T$ we define $d(C^T,x)=\min_{c \in C^T} d(c,x)$, slighting abusing notation by overloading the meaning of distance $d(\cdot,\cdot)$.
All of the variations of distance functions above then apply directly.

\subsection{Theoretical Analysis}
\label{S:theory}

Recall that an important theoretical question in our proposed approach is whether focusing on the finite-sample approximation in Problem~\eqref{E:promotion_approx} is principled, in the sense that it yields a provable approximation guarantee even vis-a-vis the true objective (Problem~\eqref{E:promotion}).
We now address this issue, focusing on promotion attacks; the analysis for demotion attacks is essentially identical.
Note that the answer is not self-evident.
First, we are not merely approximating the expectation with sample average (and removing the constant factor, which does not impact the optimal solution), but approximating an \emph{optimization problem}.
Thus, law of large numbers does not immediately imply that the quality of solution to the approximate problem converges.
Second, the problem involves bi-level optimization, since the attacker's optimal solution depends, in turn, on the learning optimization problem that yields the reward model parameters $\theta$.
Next, we prove that this problem has polynomial sample complexity, that is, the number of samples $N$ grows only polynomially in $1/\epsilon$ and the dimension of the function class $R_\theta$, where $\epsilon$ represents the quality of the approximation of the solution to Problem~\eqref{E:promotion} by the solution to Problem~\eqref{E:promotion_approx}.

\begin{lemma}
Suppose $\mathcal{F}$ is a vector space of real-valued functions, and $H = \{\mathbf{1}_{\{f + f(a) \geq 0\}}: f \in \mathcal{F}\}$, where $a$ is a constant, then VCdim$(H) \leq \text{dim}(\mathcal{F}) + 1 $, where $\text{dim}(\mathcal{F})$ is the dimension of the vector space of functions $\mathcal{F}$.
\label{vcdim}
\end{lemma}
\begin{proof}
    First, we discuss the case where $f_i(a)$ has the same sign for all $i$. 
    Let $\{f_1, f_2, \cdots, f_d\}$ be a basis of $\mathcal{F}$. Then if $\{x_1, x_2, \cdots, x_m\}$ is shattered by $H$, there are vectors $v_1, v_2, \cdots, v_{2^m}$ taking all possible sign patterns, and corresponding $w_1, w_2, \cdots, w_{2^m} \in \mathbb{R}^d$ such that $M(w_1 \cdots w_{2^m}) = V - C$, where $M_{i,j} = f_i(x_j)$, $V_{i, j} = v_{i,j}$ and $C_{i,j} = f_i(a)$.
    If $m > d$ then the matrix $M$ is not row-rank $m$. Without loss of generality, we assume the last row can be written as a linear combination of other rows, meaning for some $\alpha_1, \cdots, \alpha_{m - 1}$ we have $v_{mi} = \sum_{j = 1}^{m - 1}\alpha_j v_{ji} + (1 - \sum_{j = 1}^{m - 1} \alpha_j) \cdot f_i(a)$. If $(1 - \sum_{j = 1}^{m - 1} \alpha_j) \cdot f_i(a) \geq 0$, then we choose $i$ such that $\alpha_j v_{ji} \geq 0$ for all $j$, this means $v_{mi} \geq 0$. If $(1 - \sum_{j = 1}^{m - 1} \alpha_j) \cdot f_i(a) \leq 0$, then we choose $i$ such that $\alpha_j v_{ji} \leq 0$ for all $j$, this means $v_{mi} \leq 0$. This contradicts the assumption that $v_i$ together take on all $2^m$ sign patterns. 

    Now we discuss a general case where we do not have the assumption that $f_i(a), \forall i$ has the same sign. Let $\{f_1, f_2, \cdots, f_d\}$ be a basis of $F$. If $1 \in \{f_1, f_2, \cdots, f_d\}$, then $\{1, f_1 - c_1, \cdots, f_d - c_d\}$ is also a basis vector (they are also linearly independent), we can choose $c_1, \cdots, c_d$ such that $f_i(a) - c_i \geq 0$, otherwise, we add $1$ to the basis vectors of $\mathcal{F}$. Then we have VCdim$(H) \leq$  dim$(\mathcal{F})$ + 1.
\end{proof}

\begin{theorem}[\cite{anthony1999neural}]
Suppose that \( H \) is a set of functions from a set \( X \) to \( \{0,1\} \) and that \( H \) has finite Vapnik-Chervonenkis dimension \( d \geq 1 \). Let \( L \) be any sample error minimization algorithm for \( H \). Then \( L \) is a learning algorithm for \( H \). In particular, if \( m \geq d/2 \) then %
its sample complexity satisfies the inequality
\[
m_L(\varepsilon, \gamma) \leq m_0(\varepsilon, \gamma) = \frac{64}{\varepsilon^2} \left( 2d \ln \left( \frac{12}{\varepsilon} \right) + \ln \left( \frac{4}{\gamma} \right) \right).
\]
Here $m_L(\varepsilon, \gamma) = \min\{m : m$ is a sufficient sample size for $(\varepsilon, \gamma)$-learning  $H \text{ by } L\},
$ meaning for \( m \geq m_0(\varepsilon, \gamma) \) and $z \in Z^m$ chosen according to $P^m$,
$P^m \{ \text{er}_P(L(z)) \ge \text{opt}_P(H) - \varepsilon \} \geq 1 - \gamma.$
\label{vc_theorem}
\end{theorem}
In our result below, we capture the solution to the lower-level problem in which the reward model parameters $\theta$ are learned based on the poisoned dataset $\tilde{D}(\delta)$ as $\theta(\delta)$.
Let $\Omega$ denote the space of possible attacks $\delta$.
\begin{theorem}
\label{T:approx}
    Let $\mathcal{F} = \{R_\theta|\theta \in \Theta\}$ and %
    suppose $\text{dim}(\mathcal{F}) = d$.
    Then for all $\epsilon > 0$, and for all $N > m_0(\epsilon, \gamma)$ where $m_0(\epsilon, \gamma) = \frac{64}{\varepsilon^2} \left( 2(d + 1) \ln \left( \frac{12}{\varepsilon} \right) + \ln \left( \frac{4}{\gamma} \right) \right)$, $\max_{\delta \in \Omega} \hat{F}(\delta) \ge \max_{\delta \in \Omega} {F}(\delta) - \varepsilon $ with probability at least $1-\gamma$.    
\end{theorem}
\begin{proof}
Let $\Theta(\Omega) = \{\theta \in \arg\max_{\theta' \in \Theta} \mathcal{L}(\tilde{D}(\delta),\theta') | \delta \in \Omega\}$.
This is the set of all parameters $\theta$ of $R_\theta$ that can be induced by the adversarial label perturbations $\delta \in \Omega$.
Define $\mathcal{F}' = \{R_{\theta}(x) | \theta \in \Theta(\Omega)\}$.
Next, note that Problem~\eqref{E:promotion} is equivalent to $$\max_{\theta \in \Theta(\Omega)} \sum_{c \in C^T} \Pr_{x \sim \mathcal{P}}\{R_\theta(c) \ge R_\theta(x)\}.$$
Similarly, Problem~\eqref{E:promotion_approx} is equivalent (up to $1/N$ factor) to $$\max_{\theta \in \Theta(\Omega)} \sum_{c \in C^T} \frac{1}{N}\sum_i \mathbf{1}_{R_\theta(c) \ge R_\theta(x_i)}.$$
In this transformation, we effectively collapsed the bi-level structure into a pair single-level optimization problems, and can now make use of the machinery in Lemma \ref{vcdim} and Theorem  \ref{vc_theorem}.
Since $\mathcal{F}' \subseteq \mathcal{F}$, we have dim$(\mathcal{F}')\leq$ dim$(\mathcal{F})$. 
Let $H = \{\mathbf{1}_{\{R_\theta(x_T) - R_\theta \geq 0\}} : R_\theta \in \mathcal{F}'\}$, which is a class of functions that map from a set $X$ to $\{0,1\}$. 
Set the correct label for all $x \in X$ to be $1$ (which maximizes the attack success rate). 
Let $er_P(h) = P\{(x, y) \in Z: h(x) \neq 1\}$, and $\widehat{er}_z(h) = \frac{1}{N}|\{i: 1 \leq i\leq m \text{ and } h(x_i) \neq 1\}|$, $h \in H$. $opt_P(H) = \inf_{g \in H} er_P(g) = \max_{\delta \in \Delta} F(\delta)$, and $\widehat{er}_z(L(z)) = \min_{h \in H} \widehat{er}_z(h)$. We can then apply Lemma \ref{vcdim} and Theorem  \ref{vc_theorem} to get the desired result.
\end{proof}

%% file: experiment.tex
\section{Experiments}
\label{S:exp}

In this section we proceed with a comprehensive vulnerability analysis of reward model learning within the Bradley-Terry and MLE modeling framework described in Section~\ref{S:prelim}.

\subsection{Experiment Setup}

Our experiments involve four problem settings.
Our first experiment involves training a reward model toward safety alignment of Large Language Models (LLMs).
In the next two, reward model learning is done in the context of value alignment for control~\cite{christiano2017deep}.
The first of these generates control trajectories in MuJoCo environment associated with low-dimensional state inputs, whereas the second uses Atari to generate control trajectories with images as inputs.
The fourth considers a recommendation system context using textual information.
For this we use the Amazon rating data,\footnote{\url{cseweb.ucsd.edu/~jmcauley/datasets/amazon_v2/}, Home and Kitchen.} which we transform into pairwise comparisons.
{The motivation for exploring these four settings is to assess the performance and impact of our proposed attack for qualitatively different control problems, since the preference-based reward model learning and RLHF have been used in domains as distinct as robotic control~\cite{jain2013learning} and LLMs~\cite{ouyang2022training}.}

We conducted 5 independent runs for all but one experiment, which means that the data were collected independently for each run, and  report the mean and standard error of the mean (SEM) in the plots. The only exception is LLM, where we conducted only 1 run due to high computational requirements.
In most cases below, we learn a neural network reward model $R_\theta$, with one exception where we also investigate robustness of a linear reward model in comparison with a neural network in the same setting.
In all cases, we vary the budget $B$ between 1\% and 10\% of data.
Next we provide further details about the experiment environments and our setup; further details are provided in the Appendix.

\smallskip
\noindent\textbf{Safety Alignment:}
We use the PKU-SafeRLHF-30k dataset\footnote{\url{huggingface.co/datasets/PKU-Alignment/PKU-SafeRLHF-30K}}, which includes 30k question-answer pairs annotated with both helpfulness and harmlessness preference labels. For the scope of our study, we only use the harmlessness preference labels for our safety alignment task. Then we train the reward model initialized with an instruction following language model, which is obtained by performing supervised fine-tuning on the pre-trained model LLaMA-7B~\cite{touvron2023llama} using the Stanford Alpaca Dataset.%

\smallskip
\noindent\textbf{MuJoCo Control:} We use three MuJoCo environments: Reacher, Hopper, and Walker2D. 
We use trajectory data as inputs (outcomes), where  trajectory $\mathcal{T}=((s_1, a_1), (s_2, a_2), \cdots (s_l, a_l))$ is induced by a random policy $a_i = \pi(s_i)$, with $s_i$ representing the system state (low-dimensional and observable) and $a_i$ the action. 
Following \cite{christiano2017deep}, we record the total reward returned by the MuJoCo simulator as the reward for the trajectory. The trajectory with the higher reward is marked as winning, and the other as losing. 
The trajectory length for each environment is 30. 

\smallskip
\noindent\textbf{Atari Vision-Based Control:} For Atari control with image inputs, we experiment with three environments: Pong, Breakout, and Qbert. 
We used trajectory data as inputs as in the previous setting, except now our states $s_i$ are images.
We concatenate actions with image embeddings in the mid-linear layer. 
For experiments with PCA we reduces the input dimension to $20$, and we then concatenate it with the action. 

\smallskip
\noindent\textbf{Recommendation System:} 
Finally, we use Amazon Rating Data collected with the purpose of designing recommendation systems.
While this, like many conventional recommendation datasets, relies on user ratings, it has been observed that learning utilities using pairwise comparisons rather then relying solely on ratings can yield better recommendations~\cite{kalloori2018eliciting}.
Consequently, we used this data to generate a derived dataset of pairwise comparisons by comparing the recorded ratings of pairs of comments, which were used as inputs (outcomes).
We used BERT to generate embeddings for each input comment, considering it as the input representation. 
When we apply PCA instead, we reduce the original dimension to 20 components.
In this domain, we experiment using both a linear neural network and a 3-layer neural network with ReLU activations (multilayer perceptron (MLP)). For linear neural networks, the average training accuracy was $92.3\%$, and the average test accuracy was $91.5\%$. 
For the neural network architecutre, average training accuracy was 100\%, while average test accuracy was 92.3\%.
While it seems that the latter architecture is superior to the linear model, our experiments below offer an interesting qualification in that regard.

\subsection{Results}
We present the results below in the case of promotion and demotion attacks for a single target outcome $c^T$, with the results considering multiple target candidates deferred to the Appendix.
In promotion attacks, we choose the target outcome as the least preferred outcome in the dataset in terms of the reward model $R_\theta$ learned on clean data.
Similarly, when considering a set $C^T$, we choose the set of smallest-reward outcomes to promote.
Demotion attacks, on the other hand, aim to demote that best outcomes in terms of $R_\theta$ learned on clean data.
Consequently, our attack settings provide the most challenging attack tasks to accomplish.

\smallskip
\noindent\textbf{Safety Alignment.}
Our first consideration involves the safety alignment problem in the context of language models.
Specifically, we consider a dataset of pairwise comparisons involving relative harmfulness of prompt responses.
We learn both the clean, and poisoned reward model $R_\theta$ by fine-tuning the LLaMA-7B model using the Stanford Alpaca Dataset.
As gradient-based methods are not scalable in this setting due to the size of the neural networks, we only consider the RBD-based approaches for attacks. 
\begin{figure}[htbp]
        \centering
    \includegraphics[width=0.9\linewidth]{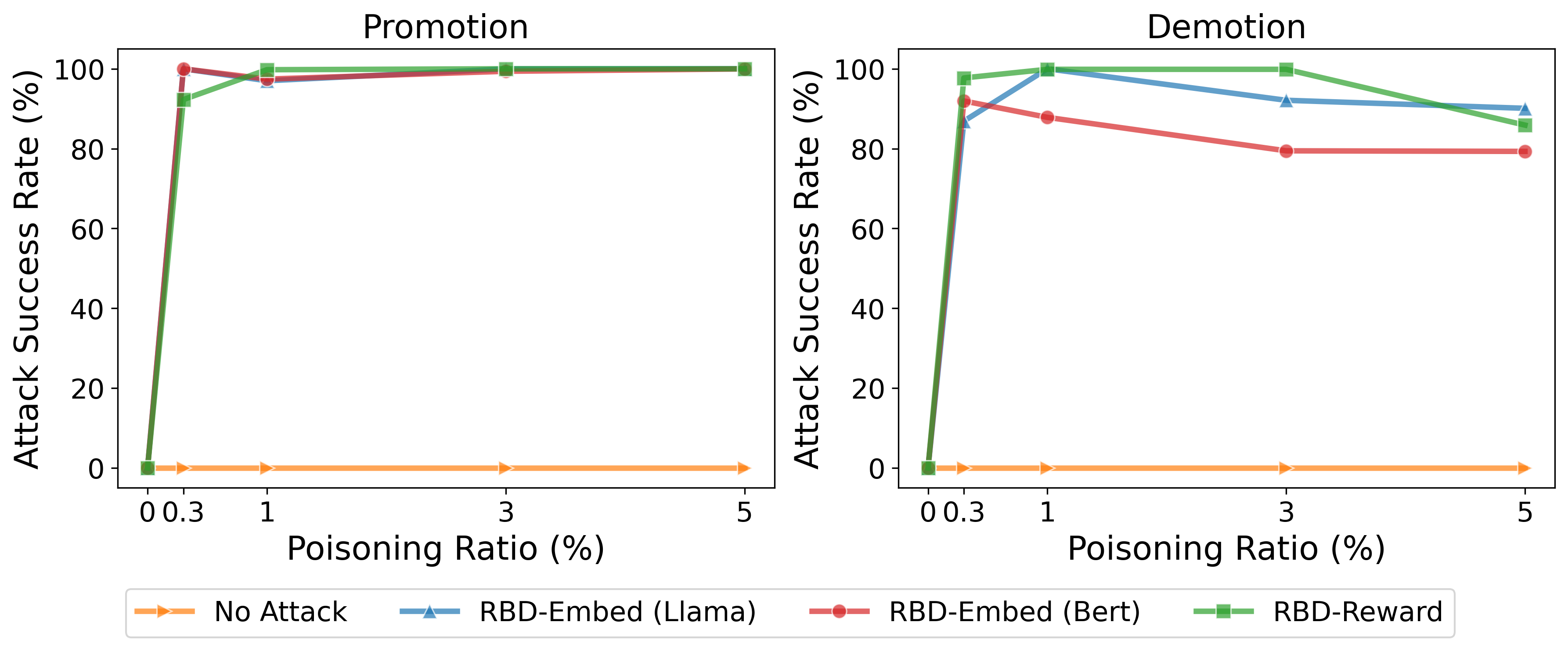}\\
    \includegraphics[width=0.9\linewidth]{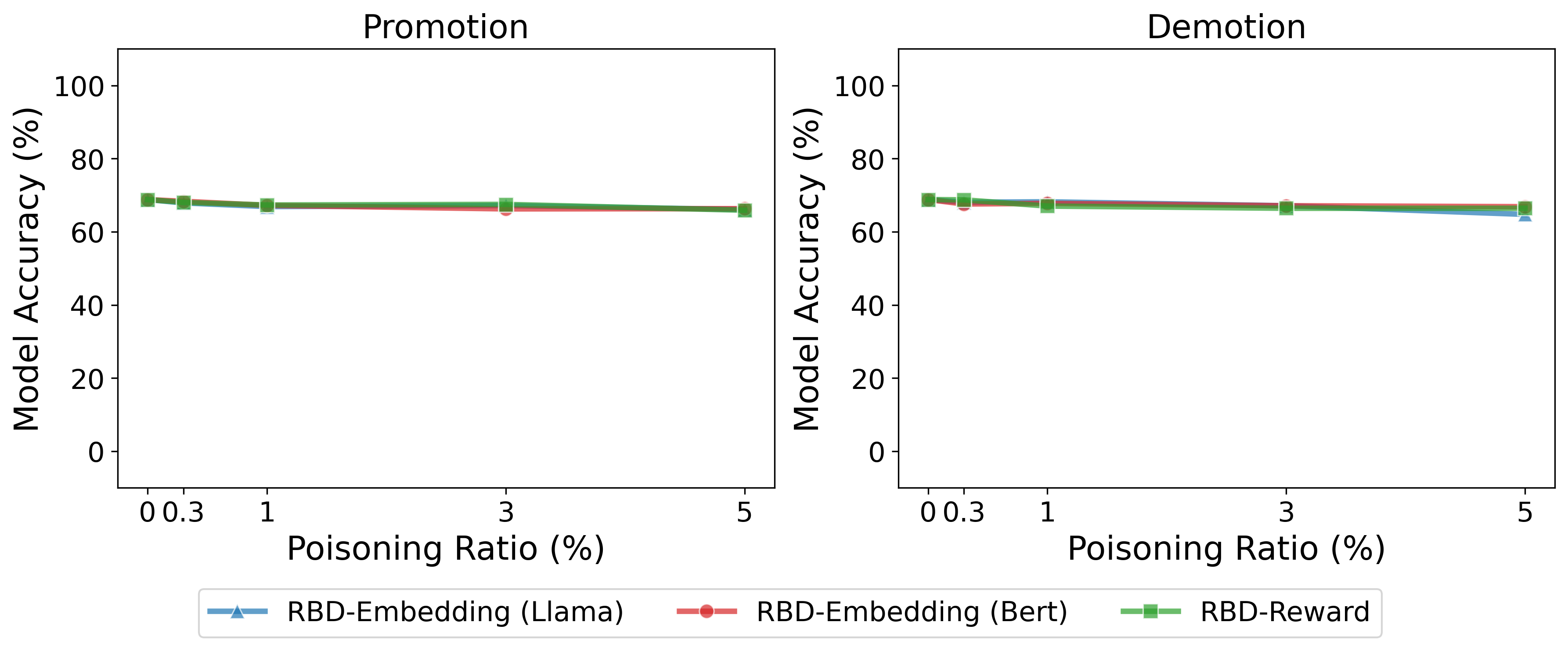}
\caption{Efficacy of promotion (left) and demotion (right) attacks in terms of success rate (top) and stealth, as measured by test accuracy (bottom) in the LLM safety alignment setting.}
\label{F:llm}
\end{figure}

Figure~\ref{F:llm} presents the results on the efficacy of the poisoning attacks (top) and accuracy degradation (bottom).
What we observe is that the LLM setting is extremely vulnerable: the best attack achieves 100\% success rate for both promotion and demotion goals with only 0.3\% poisoned instances.
A key reason is the structure of this and similar datasets: outcome comparisons are made with respect to a relatively limited set of prompts and responses, which entails many identical outcomes in pairwise comparisons.
Consequently, it is often the case that there are a large set of outcomes similar to the target, and flipping the comparison label for all of these (or any subset) is highly efficacious.
We also observe that in this setting, both RBD-Reward and RBD-Embedding (using Llama embedding) perform comparably.
In addition, we see minimal accuracy degradation as a result of the attack.

\begin{figure}[htbp]
    \centering
\includegraphics[width=\linewidth]{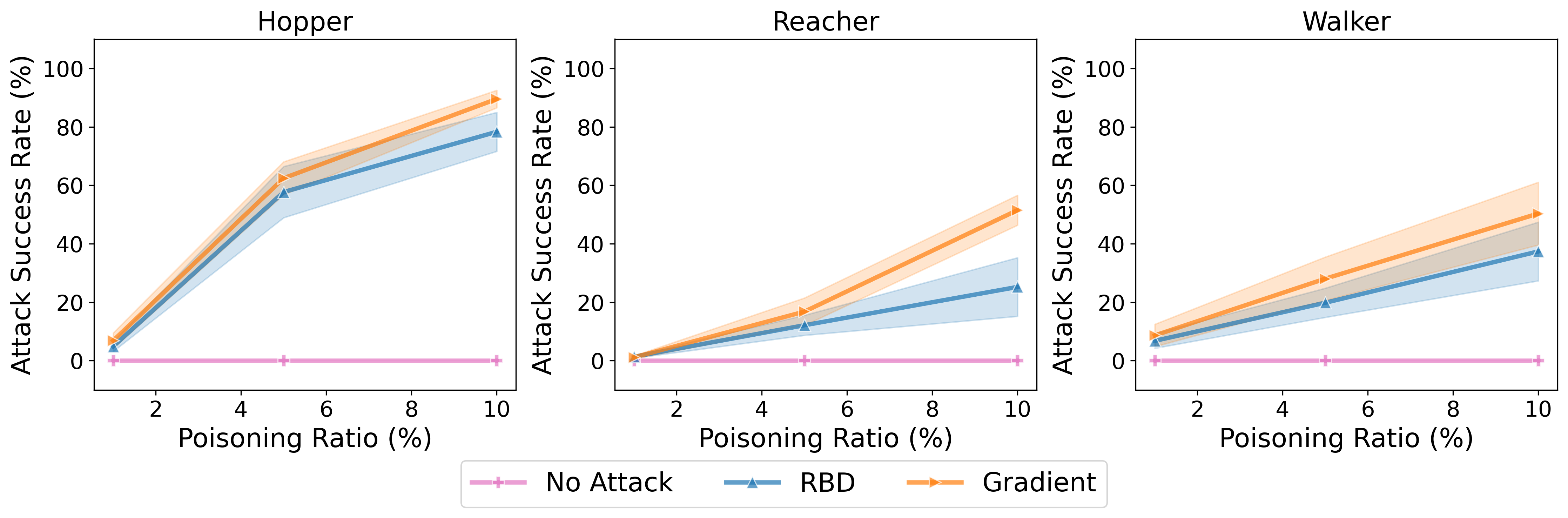}\\
\includegraphics[width=\linewidth]{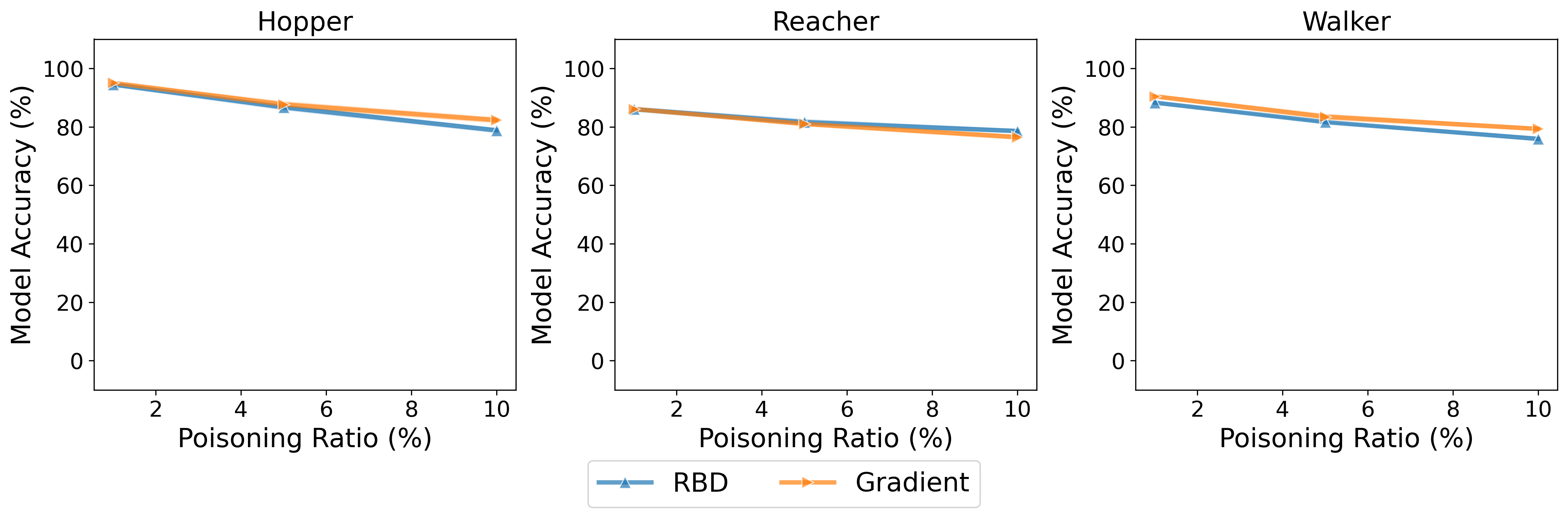}
\caption{Promotion attack efficacy in MuJoCo. Top row: success rate. Bottom row: accuracy.}
\label{F:mujoco-promotion}
\end{figure}

Next, we evaluate the downstream impact of reward model attacks on RLHF in this setting.
The success rates (of the learned policy choosing the target response to a prompt) are provided in Table~\ref{T:llm_rlhf} for a randomly chosen target candidate, for whom baseline success rate (with no attack) is $<1$\%.
\begin{table}[h]
\centering
\caption{Promotion attack efficacy in RLHF.}
\begin{tabular}{|c|c|c|c|}\hline\hline
Poisoning Ratio (\%) & 1 & 3 & 5\\\hline
Attack Success Rate (\%) & 93 & 93.5 & 99\\\hline
\end{tabular}
\label{T:llm_rlhf}
\end{table}
What we can see is that while the efficacy of the attack degrades somewhat compared to targeting reward model learning directly, it is nevertheless quite successful, achieving 99\% success rate with only 5\% of labels flipped.

\smallskip
\noindent\textbf{MuJoCo Control.}
Next we consider promotion attacks in the low-dimensional MuJoCo setting.
Figure~\ref{F:mujoco-promotion} (top) presents attack success rate for the best-performing variants of each attack class.
Our first observation, which will be echoed in other domains below, is that attack success rate (and, thus, vulnerability) varies quite significantly with problem setting.
For example, success rate in Hopper reaches over 60\% with 5\% of the attack budget, and approximately 90\% with 10\% of the budget.
On the other hand, success rate for Reacher and Walker is just over 50\% even with 10\% of the budget.

Our second observation is that in all three domains, the (best) gradient-based method is most effective.
Nevertheless, we do note that the (best) RBD heuristic is typically competitive with the best attack overall, particularly at small budgets.
For example, at 1-5\% of the budget, the performance of both classes of attacks is similar.

Next, we consider stealth of the attack with increasing attack budget (Figure~\ref{F:mujoco-promotion} (bottom)).
Although none of the methods we developed are explicitly stealthy, the results show that our targeted attacks have a limited impact on overall accuracy, particularly at the smaller range of attack budgets between 1\% and 5\%.

\begin{figure}[htbp]
    \centering
\includegraphics[width=\linewidth]{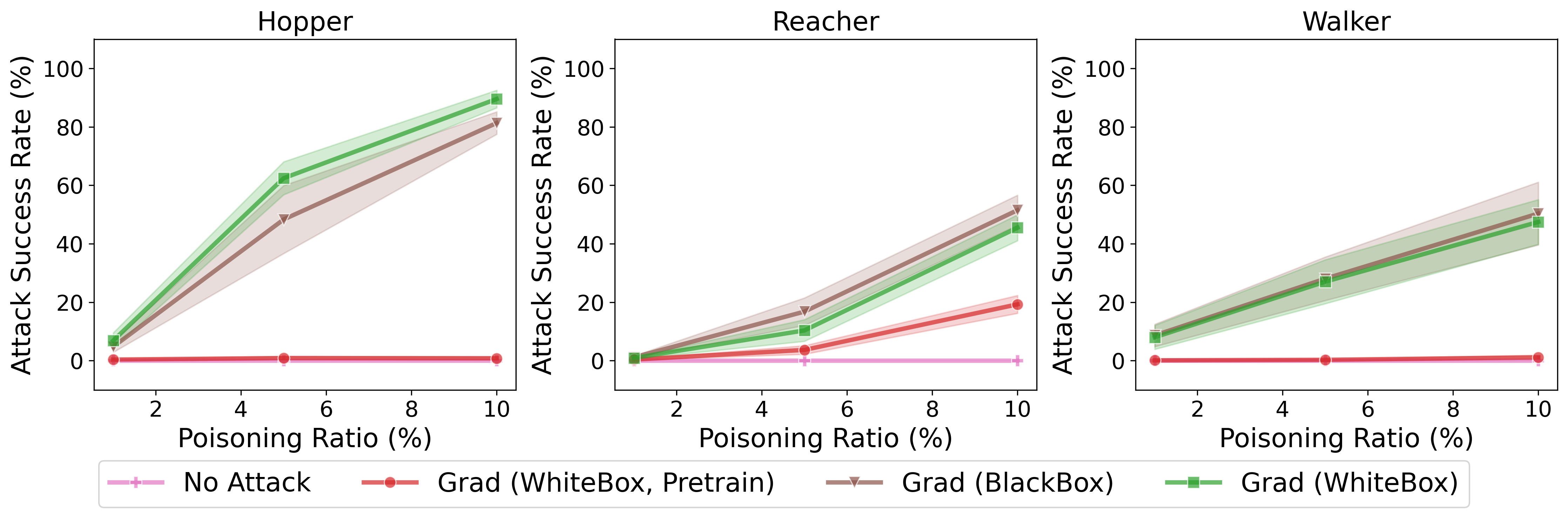}\\
\includegraphics[width=\linewidth]{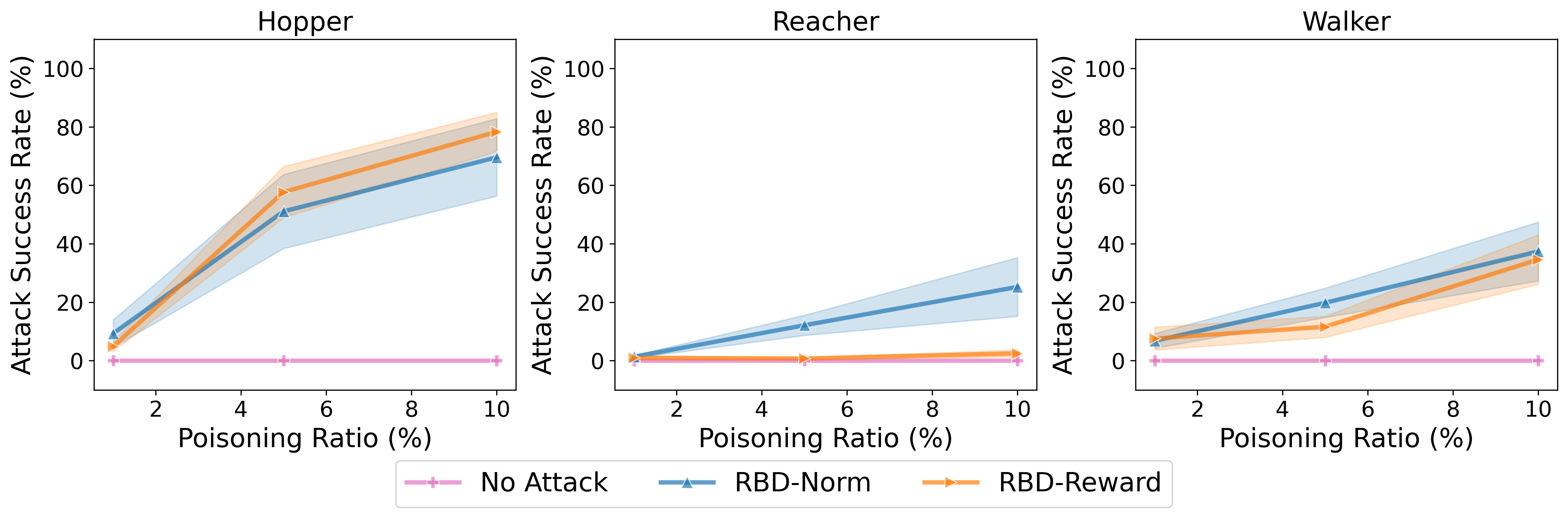}
\caption{Relative efficacy of gradient-based attacks (top row) and RBD attacks (bottom row) in MuJoCo.}
\label{F:mujoco-promotion-ablations}
\end{figure}

In Figure~\ref{F:mujoco-promotion-ablations}, we consider now ablations in terms of different gradient-based approaches (top) and RBD methods (bottom).
Here we make two observations.
First, in this setting, pretraining the neural network before performing the gradient-based attack, as done by \cite{koh2017understanding}, performs extremely poorly, yielding a nearly zero success rate.
This is likely because it is fragile to uncertainty about neural network initialization.
Rather, our primary approach in which we train the attack to be explicitly robust to initialization uncertainty performs best.
In addition, we note that black-box attacks in this context are essentially as effective as white-box attacks, suggesting that we do not need precise knowledge of the neural network architecture.
Second, we generally find that RBD-Norm outperforms RBD-Reward in two of the three environments, with the two exhibiting similar performance in the third (Hopper).

\begin{figure}[htbp]
    \centering
\includegraphics[width=\linewidth]{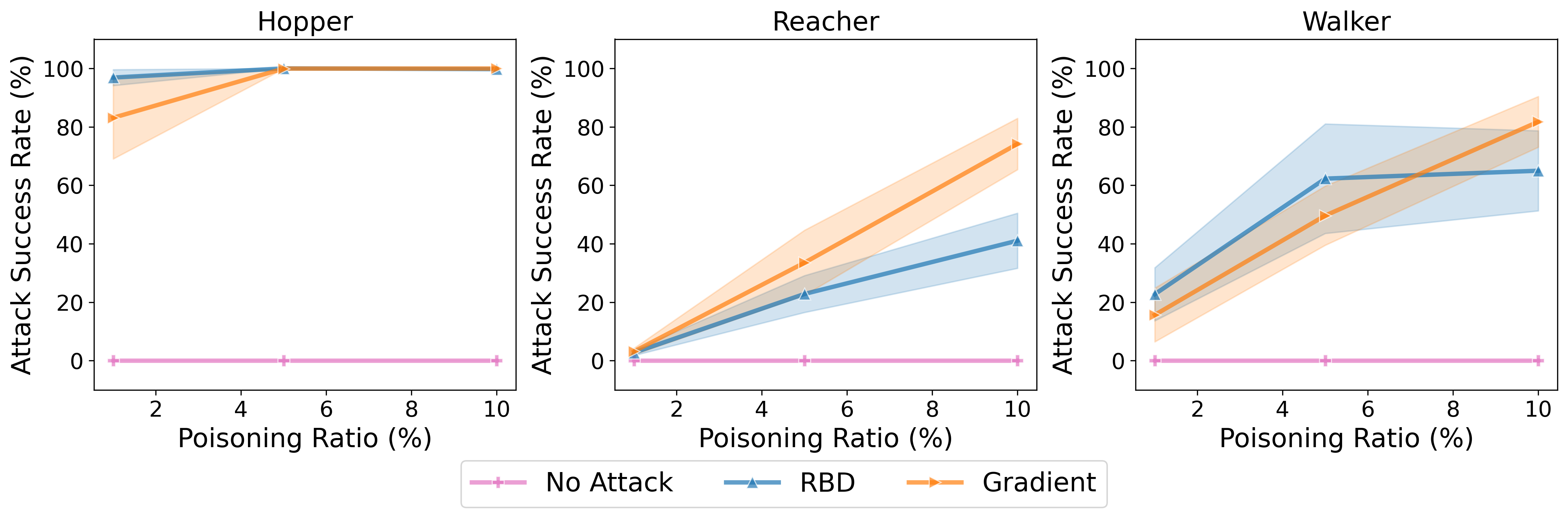}
\caption{Demotion attack efficacy in MuJoCo.}
\label{F:mujoco-demotion}
\end{figure}
Finally, Figure~\ref{F:mujoco-demotion} presents the results of demotion attacks.
We can observe that in this setting, demotion attacks are somewhat easier than promotion attacks, with success rates of the best attacks tending to be higher.
The overall trends, however, parallel what we observed in promotion attacks: gradient-based methods outperform RBD as budget increases.
On interesting exception is the demotion attack in the Hopper environment, where low-budget settings yield better RBD success rate than gradient-based method, with the former nearly 100\% successful even with a budget of 1\%.
As in the case of promotion attacks, demotion attacks are also relatively stealthy (see Figure~\ref{F:mujoco_demotion_accuracy} in the Appendix) and black-box gradient-based attacks are nearly as effective as white-box (see Figure~\ref{F:mujoco_demotion_ablation} in the Appendix).
Finally, the results are comparable when we consider a set of 5 target outcomes instead of just a single one (see Figure~\ref{F:mujoco_multiple_targets} in the Appendix).

\smallskip
\noindent\textbf{Atari Vision-Based Control.}
Next, we turn to the more challenging control setting in which true state of the controlled system is not directly observable, but is instead observed using (relatively) high-dimensional visual perception.

\begin{figure}[htbp]
    \centering    \includegraphics[width=\linewidth]{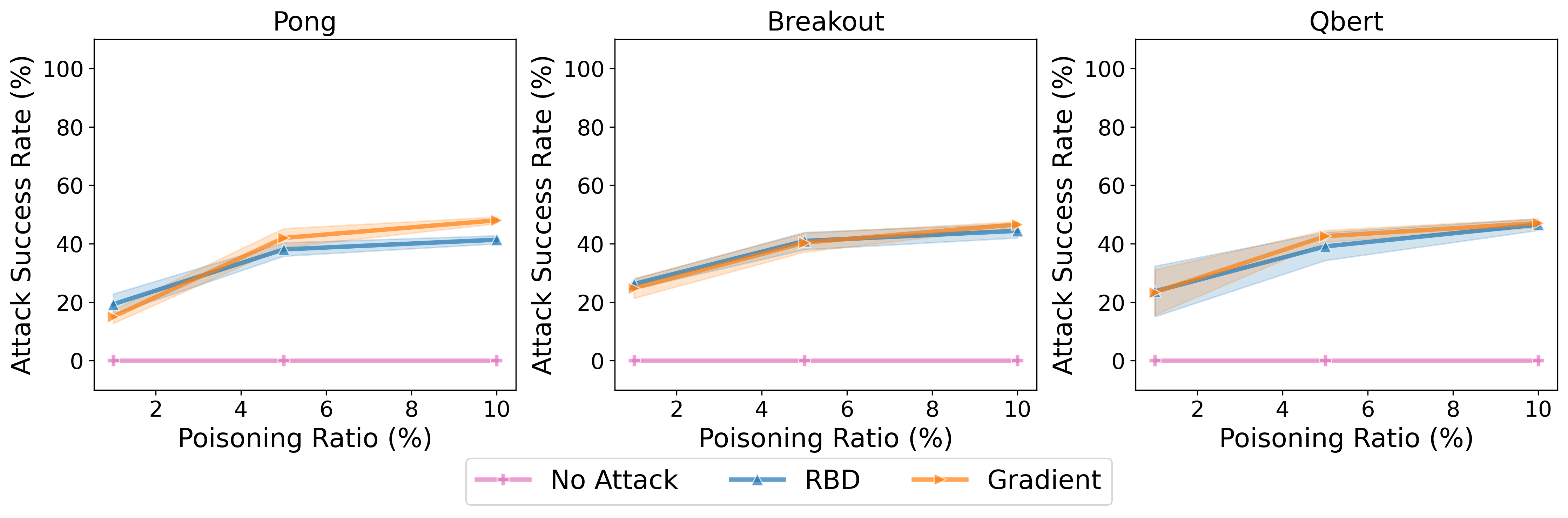}\\
\includegraphics[width=\linewidth]{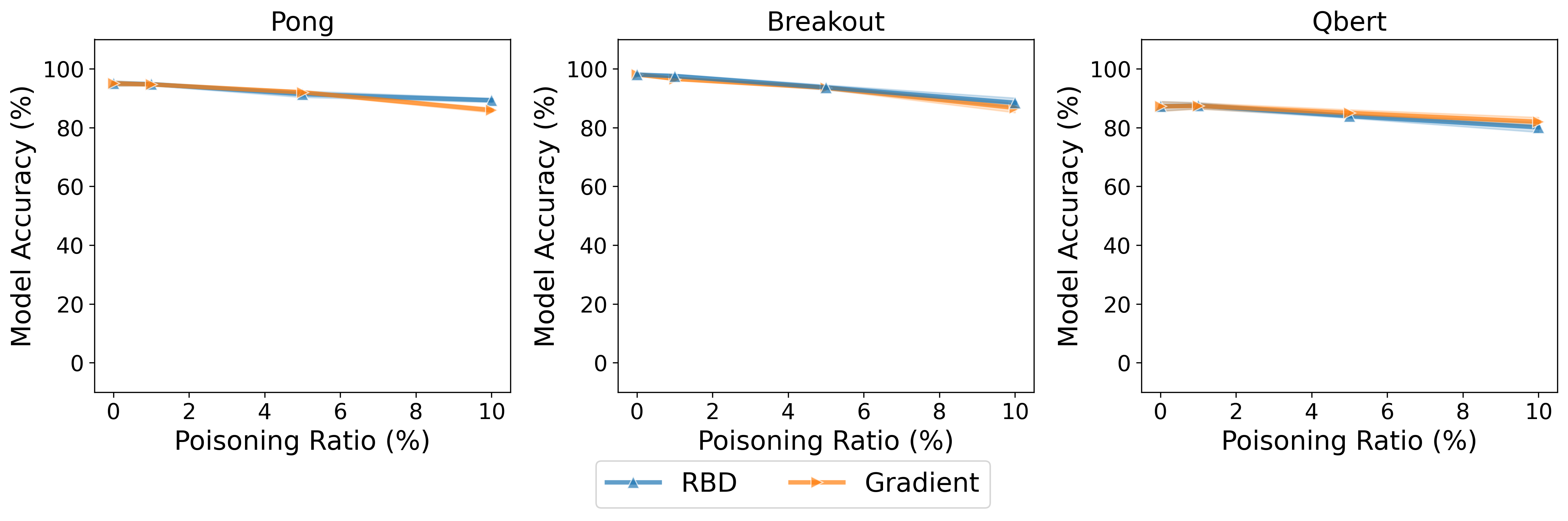}
\caption{Promotion attack efficacy in Atari. Top row: success rate. Bottom row: accuracy.}
\label{F:atari-promotion}
\end{figure}
First, as above, we consider attack efficacy in terms of both success rate (Figure~\ref{F:atari-promotion}, top) and stealth (test accuracy; Figure~\ref{F:atari-promotion}, bottom).
In this setting, success rate is markedly lower than in MuJoCo, although we still reach nearly 50\% success rate of the best attack with only 5\% of the data poisoned.
What is particularly noteworthy is that here there is little difference between the best gradient-based method and the best RBD heuristic.
As RBD is simpler, considerably faster, and requires no information about the model, this is noteworthy as one of the sources of threat are individuals themselves who are asked for their relative preferences over outcomes.
Thus, the simplicity and intuitive nature of RBD, along with its efficacy, may well make the attack a particularly significant concern in practice.

\begin{figure}[htbp]
    \centering    \includegraphics[width=\linewidth]{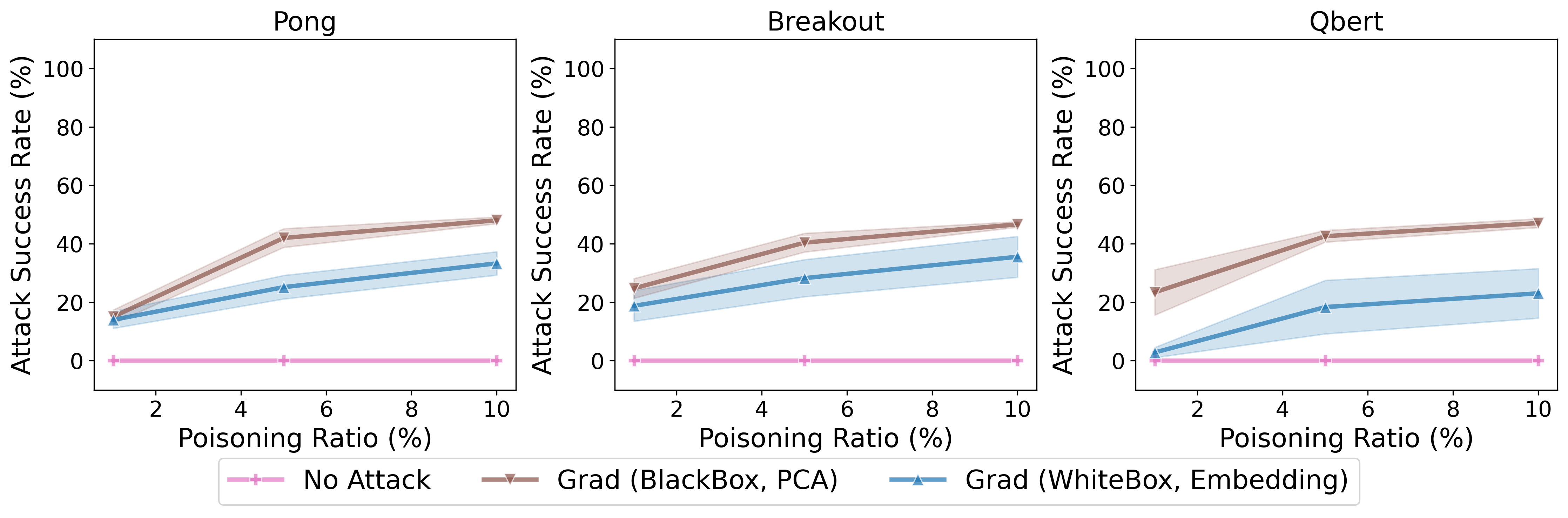}\\
    \includegraphics[width=\linewidth]{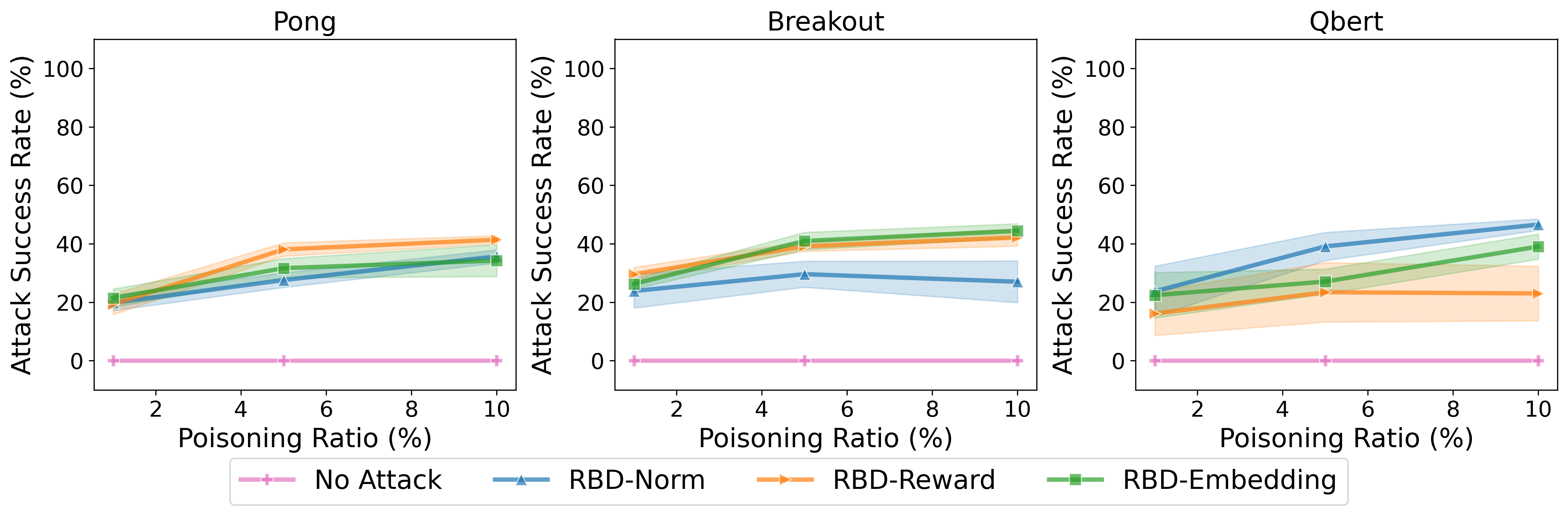}
\caption{Relative efficacy of gradient-based attacks (top row) and RBD attacks (bottom row) in Atari.}
\label{F:atari-promotion-ablations}
\end{figure}
Next, we consider again the relative efficacy of gradient-based methods (Figure~\ref{F:atari-promotion-ablations}, top row).
In this setting, surprisingly, the most effective attack is to first preprocess the high-dimensional input images using PCA (20 dimensions), and perform a (black-box) attack on a neural network over this reduced-dimensional outcome space.
This approach is significantly better than making use of the low-dimensional embedding learned as part of the reward model $R_\theta$ trained on clean data.
Additionally, 
making use of the conjugate-gradient method to approximate the inverse in implicit gradient computation, as suggested by \cite{koh2017understanding} was extremely slow, with a single attack taking $\sim$ 12 hours for one seed.

Turning now to the comparison of alternative RBD heuristics (Figure~\ref{F:atari-promotion-ablations}, bottom row), we find that their relative efficacy can now vary a great deal by environment.
In Pong, for example, all three are quite comparable, although RBD-Reward tends to outperform RBD-Embedding (which ranks by distance with respect to the embedding from $R_\theta$ learned on clean data) and RBD-Norm (which measures distances directly in pixel space).
In Breakout, RBD-Norm is the worst heuristic, while in Qbert, it is the best of the three, while RBD-Reward is the weakest one in this setting.

Our observation that relative performance of different attack methods varies considerably by domain and environment is instructive: our consideration of two classes of attack algorithms, and multiple variants in each class thereby demonstrates the value of comprehensive vulnerability analysis that this provides.
This is in contrast to common vulnerability analysis of ML to poisoning attacks in prior work, where only a single attack algorithm is typically evaluated (e.g.,~\cite{jha2023label,xiao2012adversarial,zhang2020adversarial}).

\begin{figure}[htbp]
    \centering    
    \includegraphics[width=\linewidth]{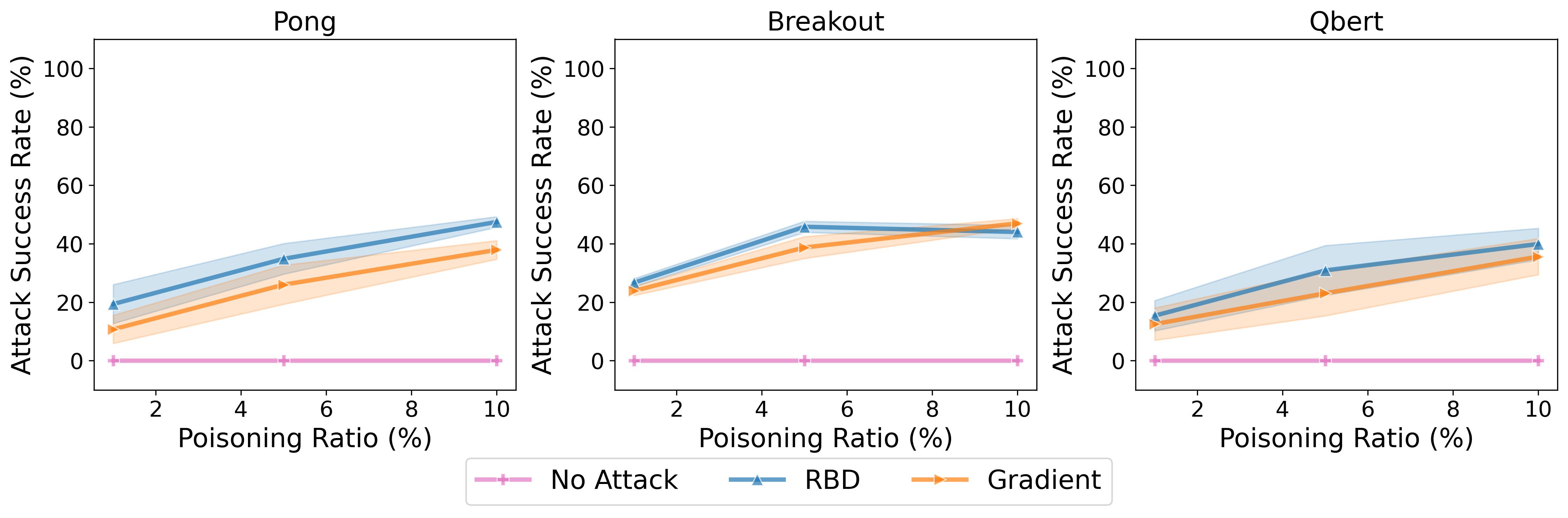}
\caption{Demotion attack efficacy in Atari.}
\label{F:atari-demotion}
\end{figure}
Finally, we consider demotion attacks in the Atari control domain.
The results are shown in Figure~\ref{F:atari-demotion}, are are broadly similar to what we have observed for promotion attacks.
However, here we see that RBD heuristics tend to outperform the best gradient-based method in nearly all cases.
This may seem surprising at first, given that the more expensive gradient-based methods are generally viewed as highly principled, but note that these are also heuristic in a number of ways.
For example, in this discrete setting, gradient-based approaches rely on relaxation of the discrete decision variables.
Moreover, as the basic variant of these fails to scale to the higher-input-dimension problem posed in vision-based control, we now also rely on the linear (PCA) dimensionality reduction technique blended with the gradient-based method (which does outperform a non-linear learned embedding).
In any case, this again reinforces our broader observation that the nature of the best attack can vary by domain, environment, and attacker's objective.

\begin{figure}[htbp]
    \centering
    \includegraphics[width=0.9\linewidth]{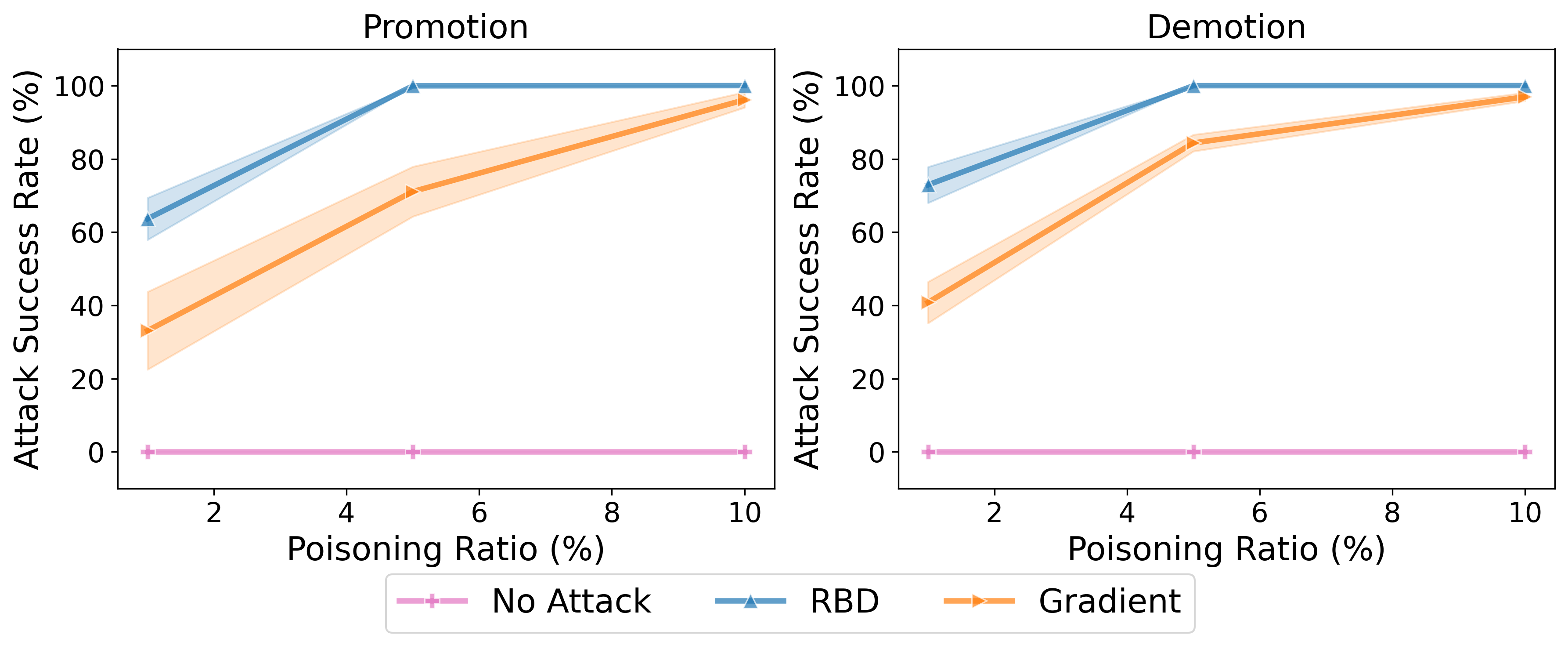}
\caption{Efficacy of promotion (left) and demotion (right) attacks on the recommendation dataset (neural network model).}
\label{F:amazon_nn}
\end{figure}

\smallskip
\noindent\textbf{Recommendation System.}
Finally, we consider a recommendation system setting.
In Figure~\ref{F:amazon_nn} we show the results of poisoning in this context for both promotion and demotion attacks, when we learn a neural network reward model (MLP).
Notably, in this setting, which is also too high-dimensional for a direct application of the gradient-based approach, RBD methods are significantly more effective than gradient-based algorithms, both in the case of promotion and demotion attacks.
Additionally, the best method (RBD) in this setting achieves 100\% success rate in both types of attacks with only a 5\% attack budget, and 60-70\% success rate with a mere 1\% attack budget.

\begin{figure}[htbp]
    \centering
    \includegraphics[width=0.9\linewidth]{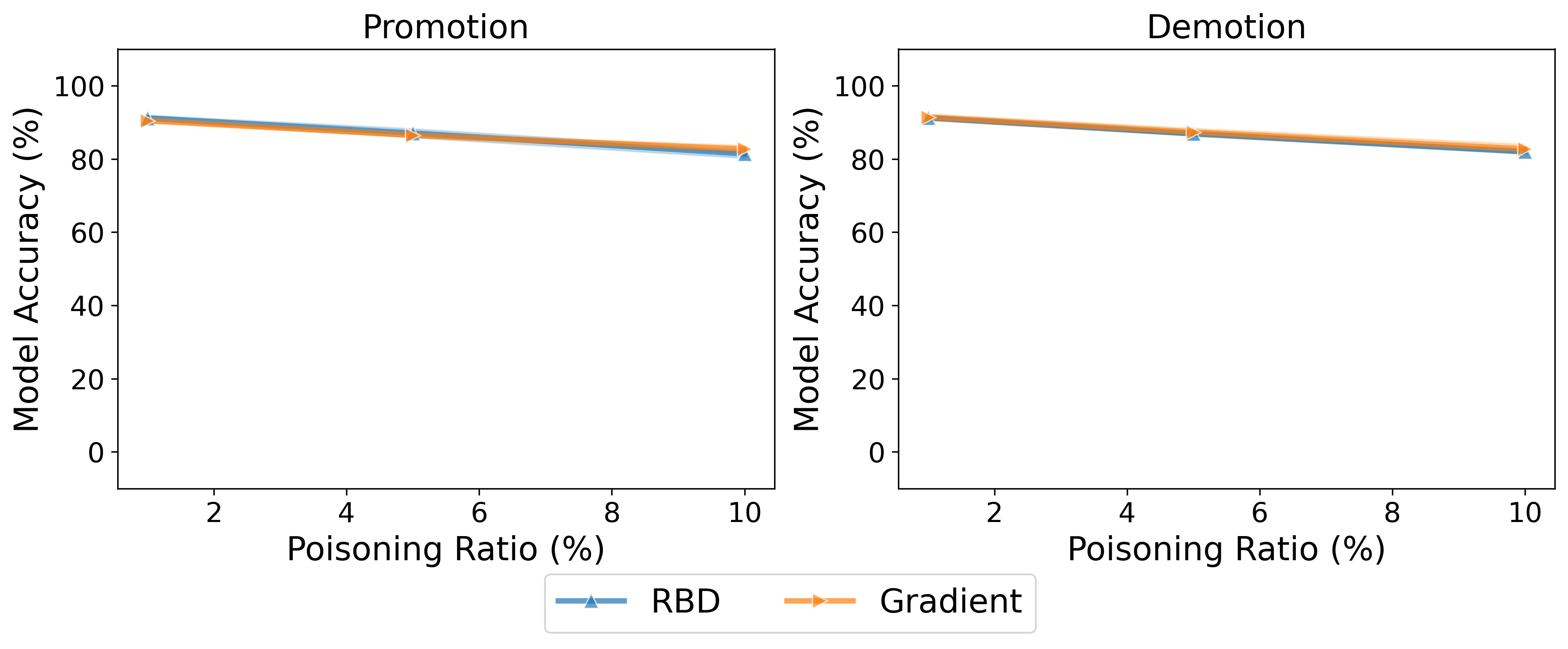}
\caption{Accuracy of promotion and demotion attacks on the recommendation dataset (neural network model).
}
\label{F:amazon_nn_accuracy}
\end{figure}

In Figure~\ref{F:amazon_nn_accuracy} we present accuracy results as a function of the attacker budget.
As in all the settings so far, our targeted attacks have only a small impact on test accuracy, both for promotion and demotion attacks.

\begin{figure}[htbp]
    \centering
    \includegraphics[width=0.9\linewidth]{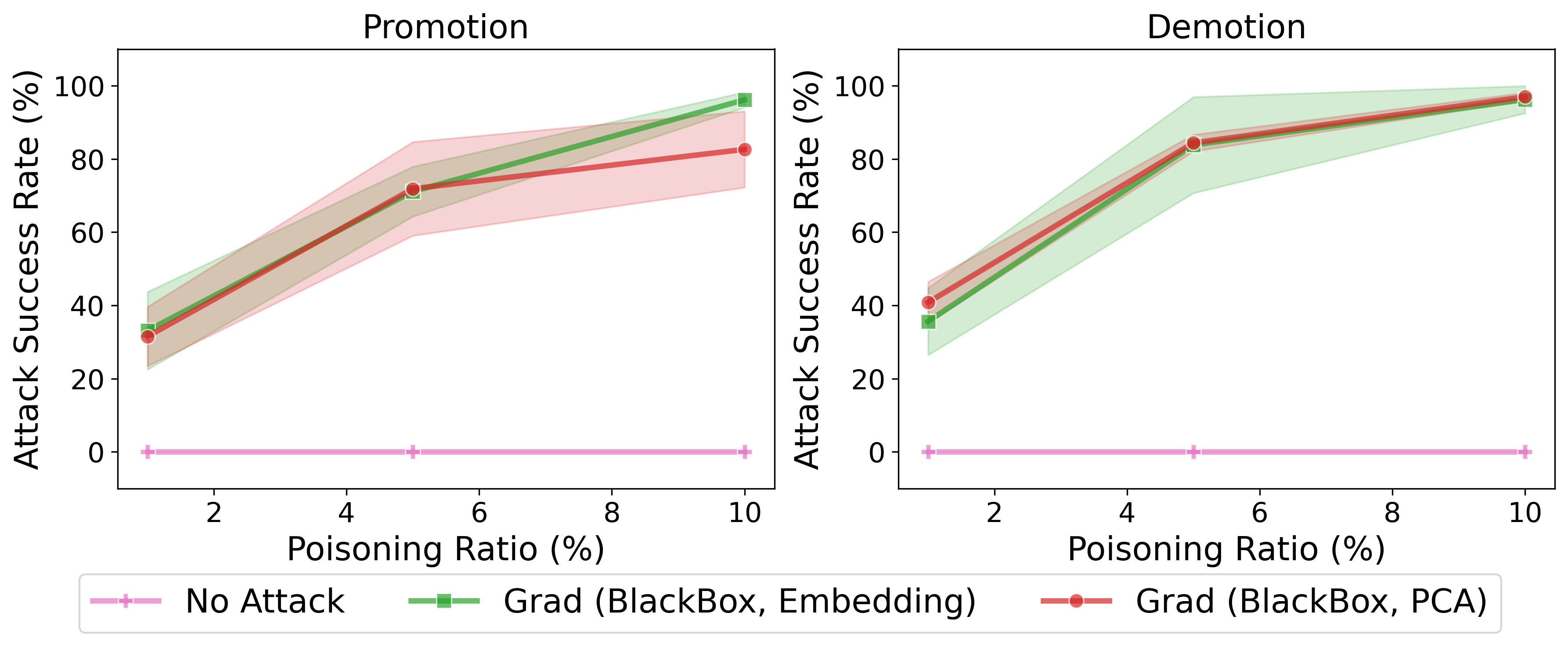}\\
    \includegraphics[width=0.9\linewidth]{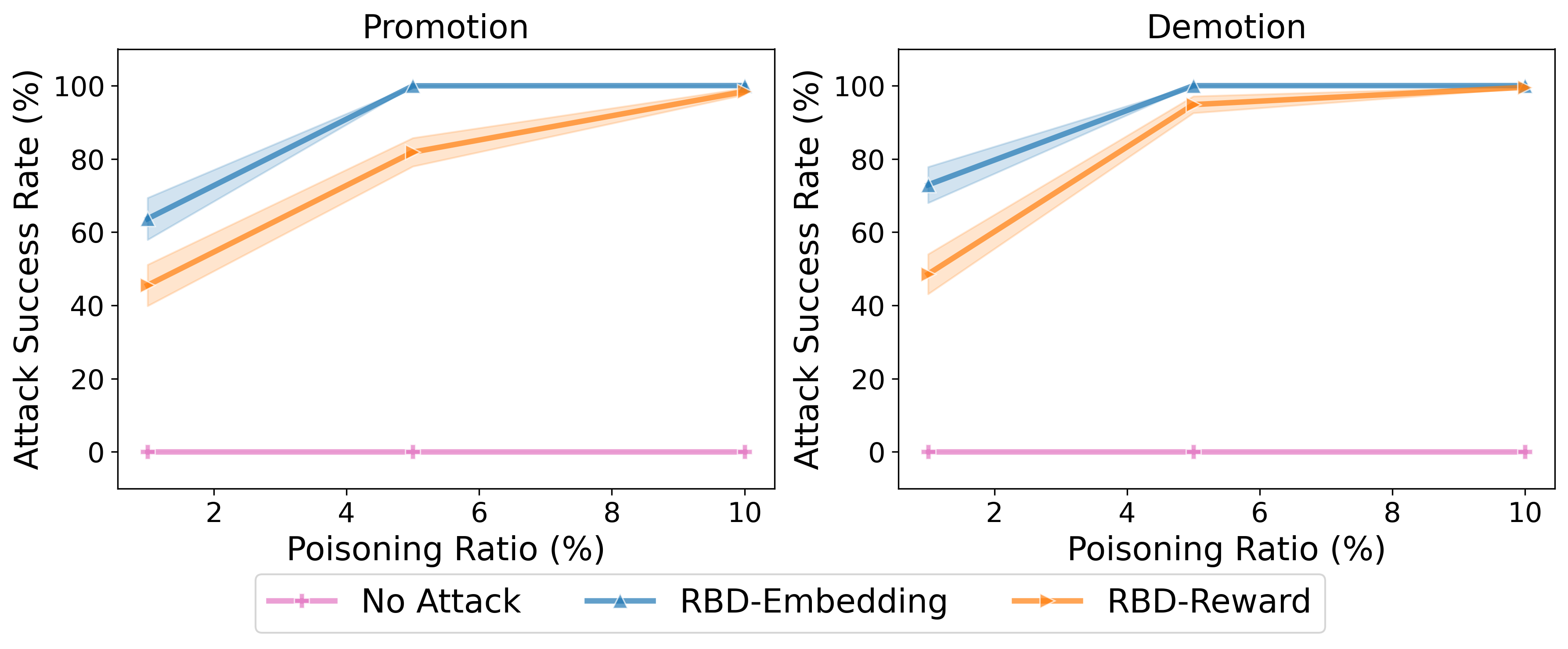}
\caption{Relative efficacy of gradient-based attacks (top) and RBD attacks (bottom) on the recommendation dataset (neural network model).}
\label{F:amazon_nn_efficacy}
\end{figure}

Consider now Figure~\ref{F:amazon_nn_accuracy}, which evaluates relative efficacy of gradient-based approaches (top) and relative efficacy of RBG approaches (bottom).
In this domain, we observe that gradient-based approaches aimed at improving scalability via the use of conjugate gradient methods still do not scale well (running a single attack takes $\sim 5$ hours), and simple PCA to reduce the input dimension can relatively effectively combine with a gradient-based approach.
Nevertheless, as we highlight above, scalability becomes a practical issue for such approaches, with the RBG heuristics now appearing to be a better option.
In the case of RBG variants, on the other hand, there is in this domain a clear advantage for using feature (BERT embedding) distance directly, rather than reward as a measure of distance.

\begin{figure}[htbp]
    \centering
    \includegraphics[width=0.9\linewidth]{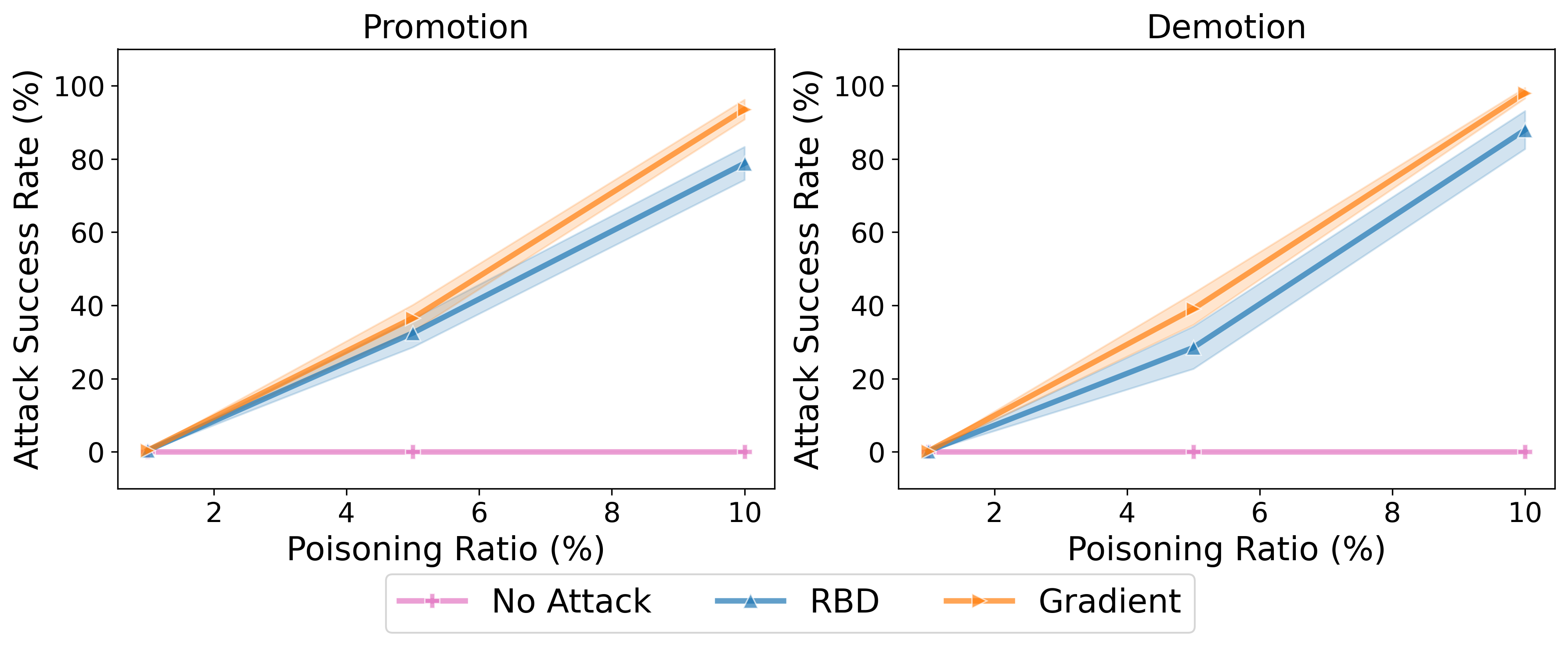}
\caption{Efficacy of promotion and demotion attacks on the recommendation dataset (linear model).}
\label{F:amazon_linear}
\end{figure}

Thus far, our consideration of reward model poisoning was focused on neural network reward models.
We now explore the extent to which using linear models %
yields better or worse robustness to poisoning attacks.
\begin{figure}[htbp]
    \centering
    \includegraphics[width=0.9\linewidth]{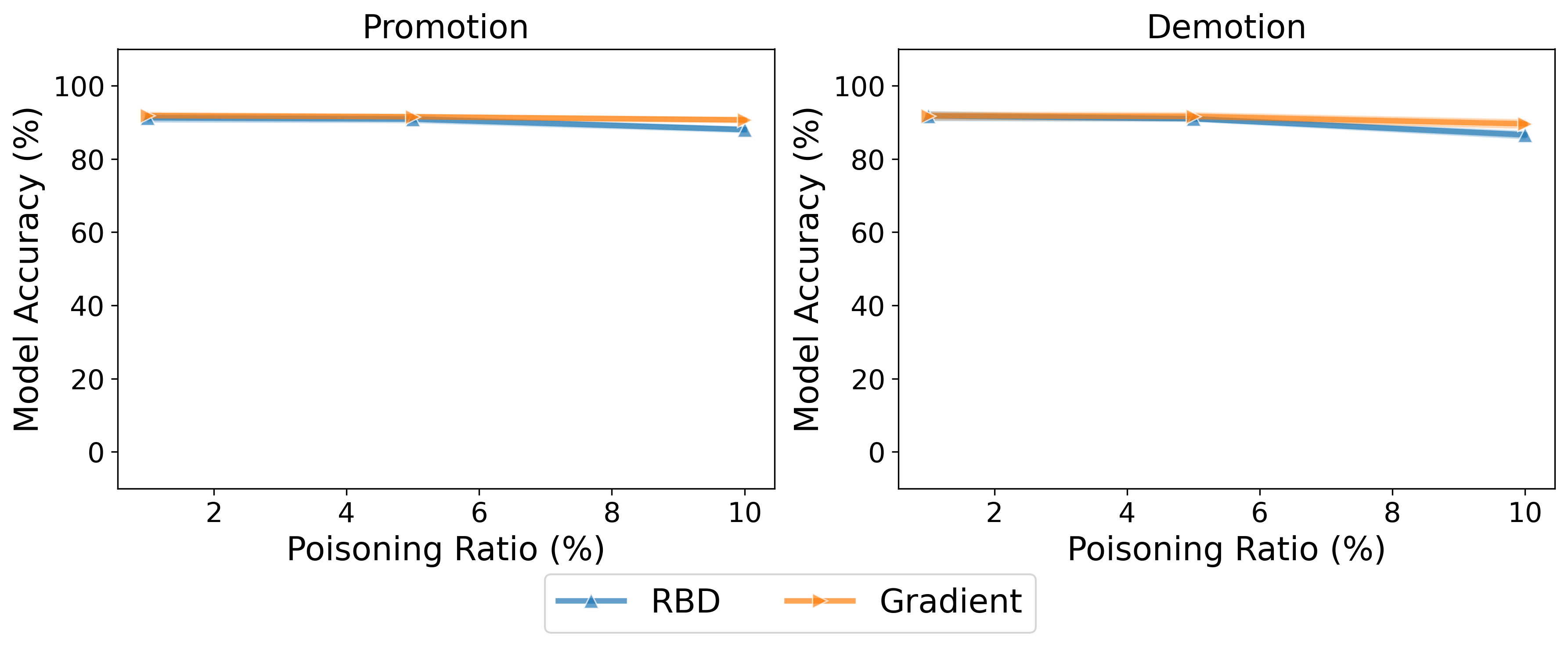}

\caption{Accuracy of promotion and demotion attacks on the recommendation dataset (linear model).
}
\label{F:amazon_linear_accuracy}
\end{figure}
In Figure~\ref{F:amazon_linear} we present the results of poisoning attacks on linear reward models.
It is instructive to compare these results to what we saw in the case of neural network reward models in Figure~\ref{F:amazon_nn}: the linear model is considerably more robust than the neural network, and it takes twice as much budget (10\%, as compared to 5\%) to reach 100\% attack success rate.
Nevertheless, both model classes are clearly vulnerable to poisoning.
The second observation we can make in the comparison is that in the case of linear models, gradient-based methods are somewhat better than RBD, whereas the reverse was true, as we noted above, in the case of neural networks.

Figure~\ref{F:amazon_linear_accuracy} presents linear model accuracy as a function of attack strength.
Here, again, we observe that accuracy degradation is minimal, and the attack is quite stealthy.

\begin{figure}[htbp]
    \centering
    \includegraphics[width=0.9\linewidth]{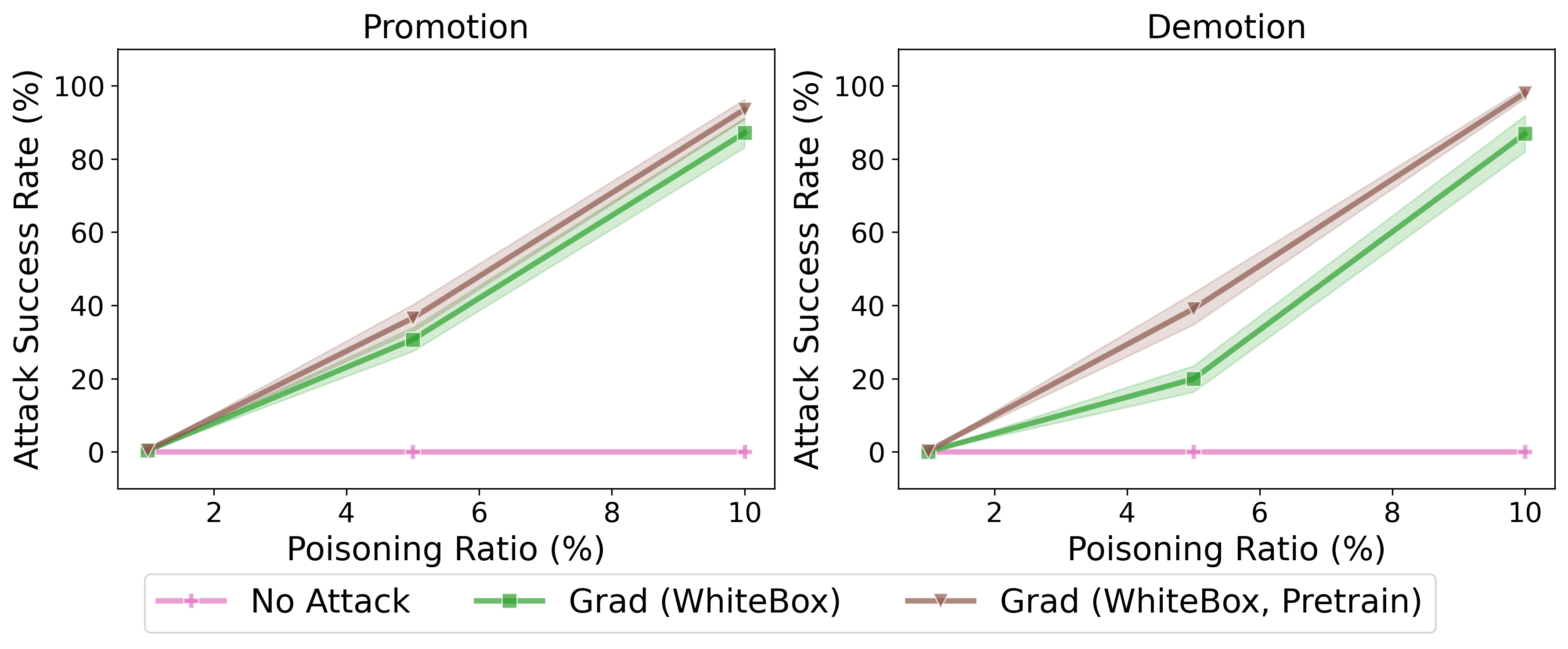}\\
    \includegraphics[width=0.9\linewidth]{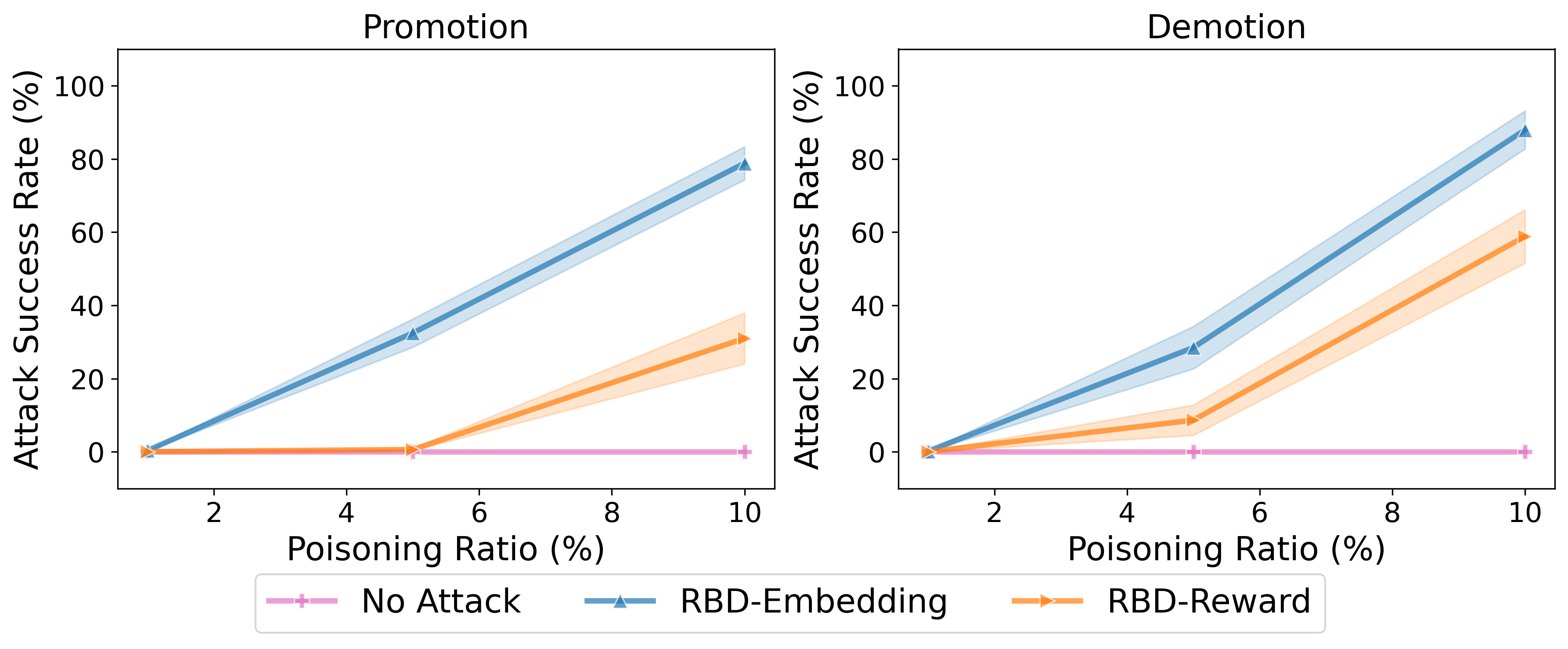}
\caption{Relative efficacy of gradient-based attacks (top) and RBD attacks (bottom) on the recommendation dataset (linear model).
}
\label{F:amazon_linear_efficacy}
\end{figure}
Finally, Figure~\ref{F:amazon_linear_efficacy} presents  ablation results comparing different gradient (top) and RBD (bottom) approaches in terms of relative efficacy.
In this domain, we find that
using a pre-trained model as part of the gradient-based scheme now outperforms the approach that instead uses $K$ random initializations, especially in the demotion attack.
In RBD comparison, on the other hand, RBD-Norm handily outperforms RBD-Reward in this setting, comparable to our observations above.

\begin{figure}[htbp]
    \centering
    \includegraphics[width=\linewidth]{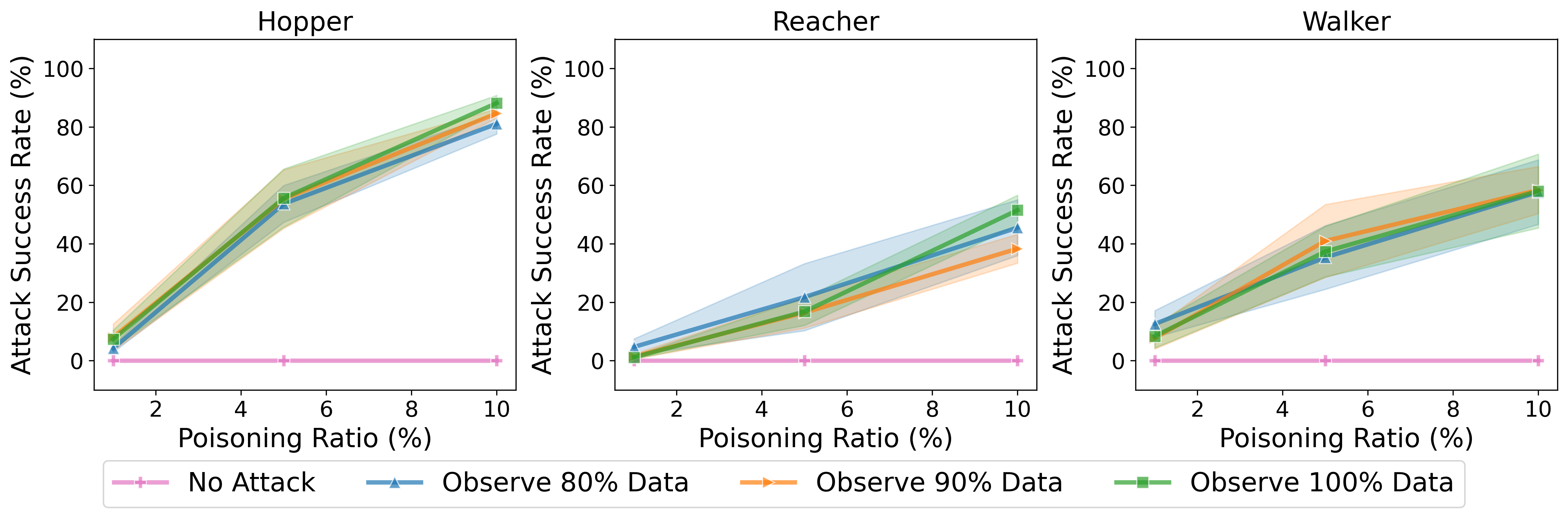}
    \includegraphics[width=1\linewidth]{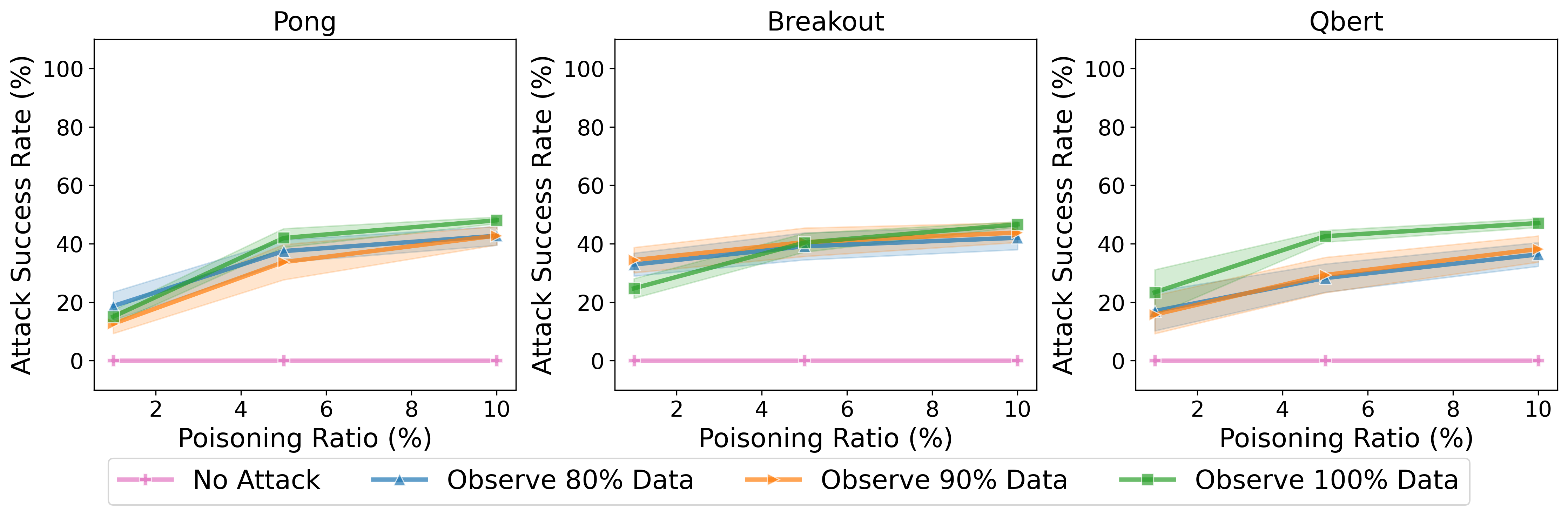}
    \caption{Promotion attack with partial knowledge of data (best attack results). MuJoCo (top) and Atari (bottom).}
    \label{fig:bbattack}
\end{figure}

\smallskip
\noindent\textbf{Partial Knowledge of Data.}
Our experiments thus far have assumed that the attacker observes the entire dataset $D$.
We now consider the impact of relaxing this assumption, when only partial information about $D$ is available.
The results are provided in Figure~\ref{fig:bbattack} for the MuJoCo and Atari control environments, and demonstrate that while attack success rate degrades under partial knowledge of data, it does so relatively slowly.

%% file: defense.tex
\section{Effectiveness of Defense} 
\label{S:defense}

There have been a number of approaches for defending against poisoning attacks.
However, these have been developed and evaluated in the context of traditional classification or regression.
Moreover, many focus on input (rather than label) poisoning, or specifically trojan attacks, whereas our problem is analogous label-flipping in classification.

We study the efficacy of several state-of-the-art defense approaches in the context of poisoning attacks on reward model learning.
Specifically, we consider representative defenses from three general classes of approaches (that can be applied in our setting as well) to defend against poisoning attacks: 1) identification and removal of anomalous (poisoned) data, 2) iterative training and removal of high-loss data, and 3) data randomization.
As representatives of the first class, we evaluate spectral anomaly detection defense, which arises from the theoretically grounded robust learning literature~\cite{tran2018a}, and the recently proposed Meta-Sift approach~\cite{zeng2023meta}.
In both \emph{spectral outlier} and Meta-Sift defense, we first apply the respective approach to identify and remove outliers, 
and then train the reward model on the rest of the training data.
Our representative of the second class is the following three-stage algorithm~\cite{du2019robust,liu2017robust,vorobeychik2018adversarial}: 1) train an initial model $R_{\theta_1}$, 2) remove $\alpha \cdot B$ datapoints with the largest loss $\mathcal{L}(\tilde{D},\theta_1)$, where $\alpha \ge 1$ is a hyperparameter (in our experiments, $\alpha=1.5$), and 3) train an updated model on the remaining data.
We refer to this as the \emph{loss outlier} defense.
Finally, our representative of the third class of approaches is the \emph{ALIBI} algorithm that leverages differential privacy and Bayesian post-processing to increase robustness to poisoning attacks~\cite{malek2021antipodes}. 

For defense strategies with ReLU neural network structure, we present results for RBD-Norm, since it is the most effective attack method. 
For defense strategies with a linear neural network structure (on the recommendation dataset), we experiment on the gradient-based approach with the pre-trained model, since it is the most effective attack method in that context.
When applying spectral signature defense in the context of text-based inputs, we use input text embedding to identify outliers, since it does not have any mid-layer embeddings.
We then arrange data such that the winning outcome is always the first, concatenate each pair of outcomes $(x_i,y_i)$, and apply the spectral signature approach to identify anomalous datapoints.

\smallskip
\noindent\textbf{Safety Alignment.}
First, we evaluate the efficacy of defense approaches in the case of safety alignment for LLM.
We exclude Meta-Sift in this setting, which requires iterative training and evaluation of the model, as it is impractical at this scale.
\begin{figure}[htbp]
    \centering
    \includegraphics[width=0.9\linewidth]{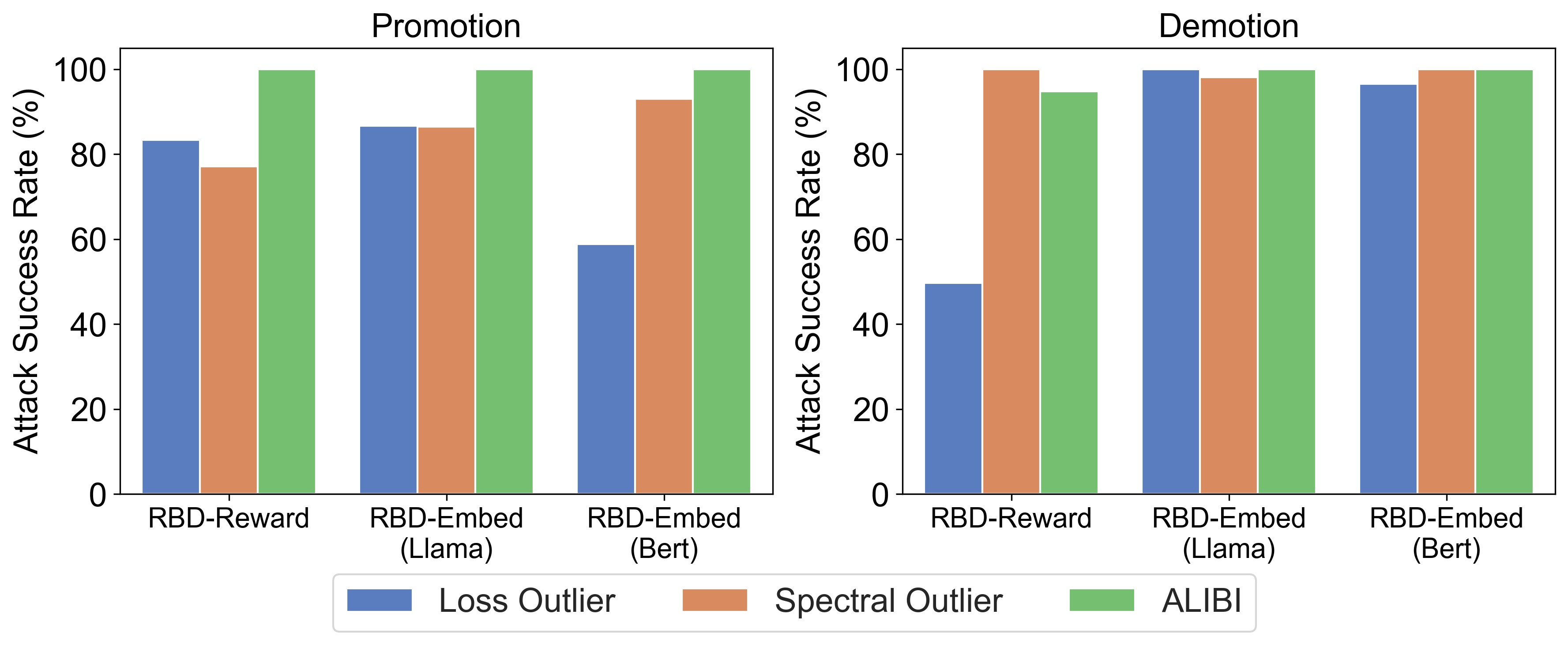}
\caption{Efficacy of defense against promotion (left) and demotion (right) attacks in the safety alignment (LLM) setting. $B=$ 0.3\% of data.}
\label{F:llm_defense}
\end{figure}
The results are shown in Figure~\ref{F:llm_defense}.
Here, in contrast with all of the other domains, none of the defensive approaches have a significant impact on the efficacy of the best attacks, even though only RBD approaches scale to this setting.
In particular, while ALIBI exhibited some success in the other settings, it is entirely ineffective even with the attack budget of only 0.3\% of datapoints being poisoned.%

\smallskip
\noindent\textbf{MuJoCo Control.} 
Next, we consider the MuJoCo setting.
\begin{figure}[htbp]
\centering
\includegraphics[width=\linewidth]{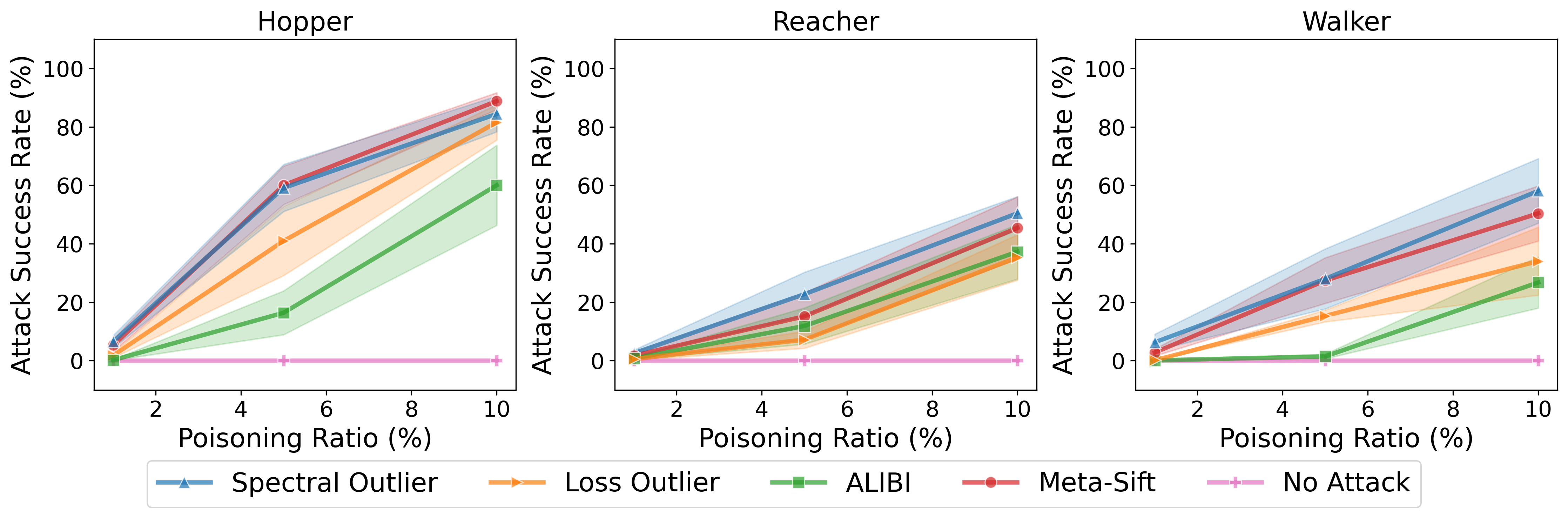}
\includegraphics[width=\linewidth]{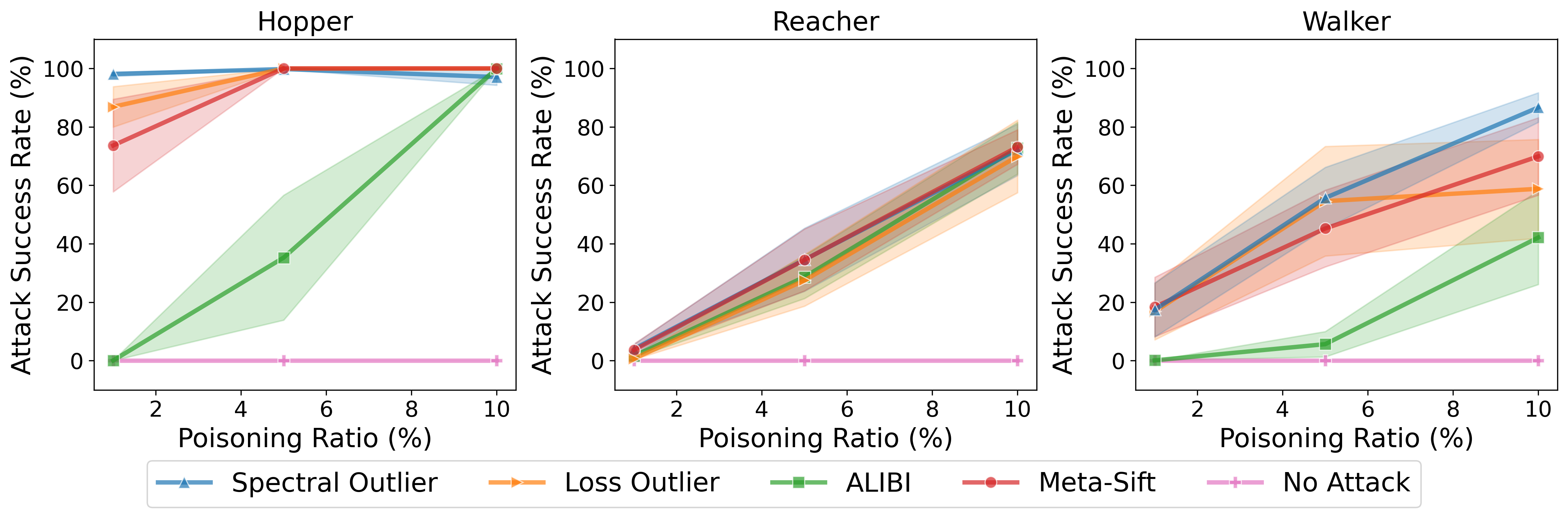}
\caption{Efficacy of defense against promotion attacks (top) and demotion attacks (bottom) in the MuJoCo setting.}
\label{F:mujoco-promotion-defense}
\end{figure}
The results are shown in Figure~\ref{F:mujoco-promotion-defense} for promotion attacks (top) and 
demotion attacks (bottom).
In most cases, we observe that neither the spectral outlier defense nor the loss outlier defense have a significant impact on the attack efficacy.
In contrast, ALIBI is considerably more efficacious in both Hopper and Walker environments, although not in the Reacher environment.
Nevertheless, with 10\% of the data poisoned, attacks remain relatively successful against all of the defenses.

\smallskip
\noindent\textbf{Atari Vision-Based Control.}
Next, we consider the efficacy of defensive methods against the proposed attacks in the Atari vision-based control domain.

\begin{figure}[htbp]
    \centering
    \includegraphics[width=\linewidth]{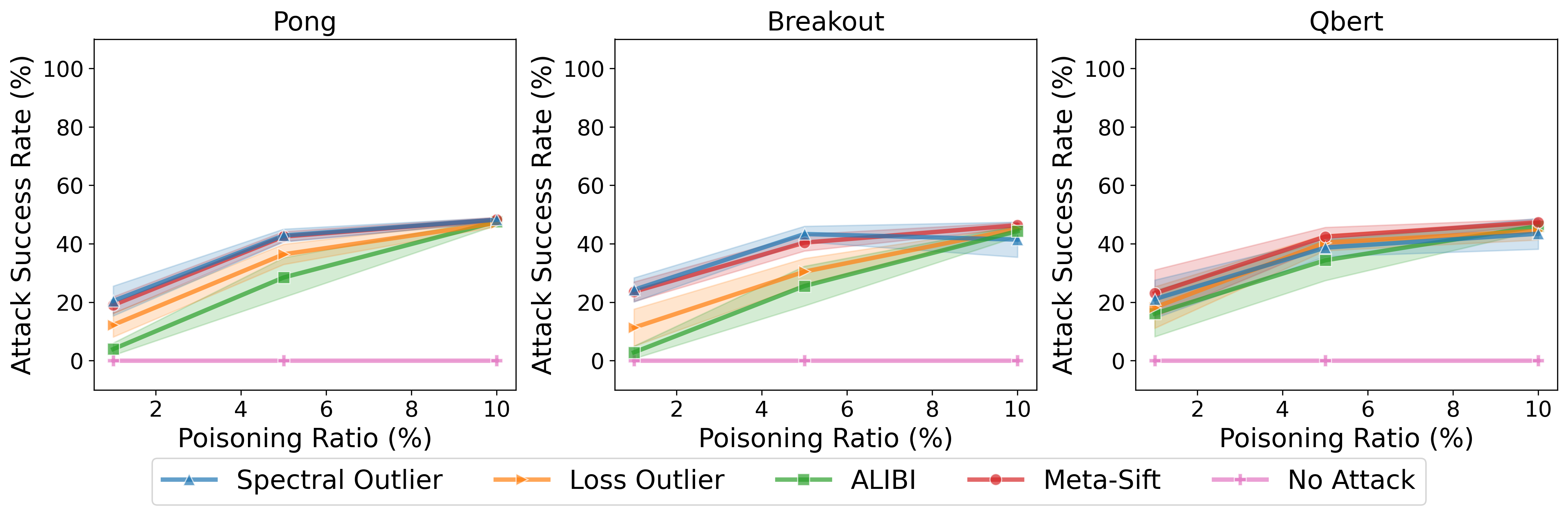}\\
    \includegraphics[width=\linewidth]{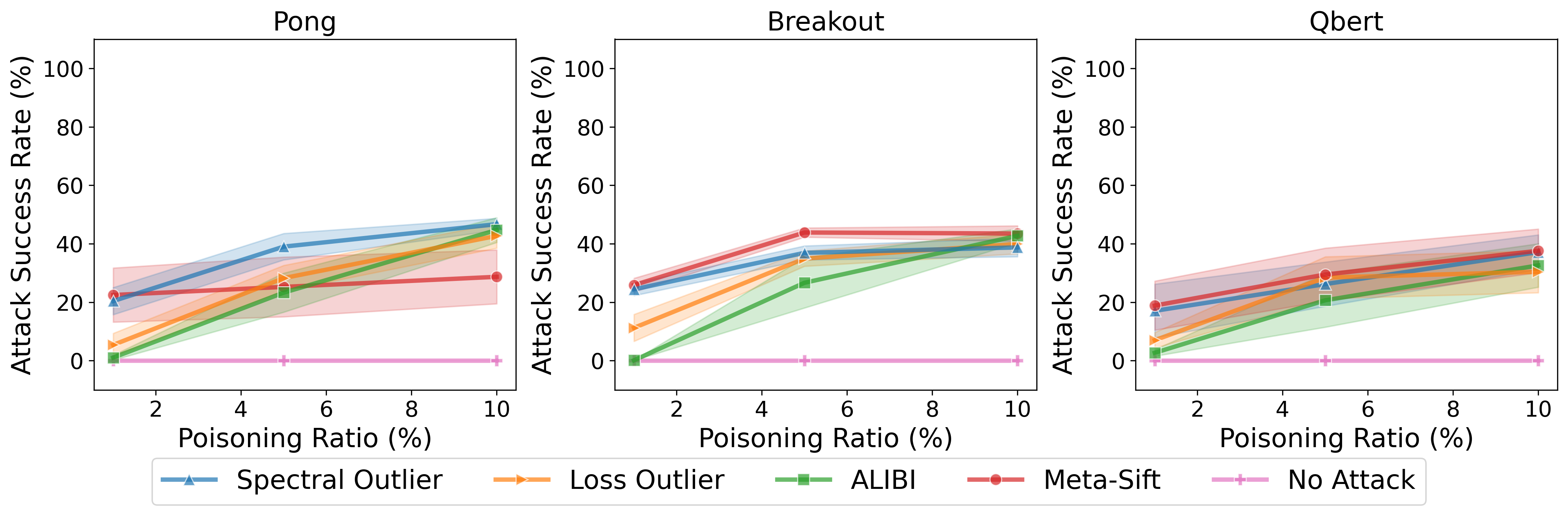}
    \caption{Efficacy of defense against promotion attack (top row) and demotion attacks (bottom row) in the Atari setting.}
\label{F:atari_defense}
\end{figure}
The results are provided in Figures~\ref{F:atari_defense}.
In this setting, the defenses again have somewhat limited impact, particularly as the attack budget reaches 10\% of the data.
While here the ALIBI defense is again consistently most effective, the loss outlier defense is typically comparable, whereas the spectral outlier is the worst of the three, and generally ineffective.

\begin{figure}[htbp]
    \centering
    \includegraphics[width=0.9\linewidth]{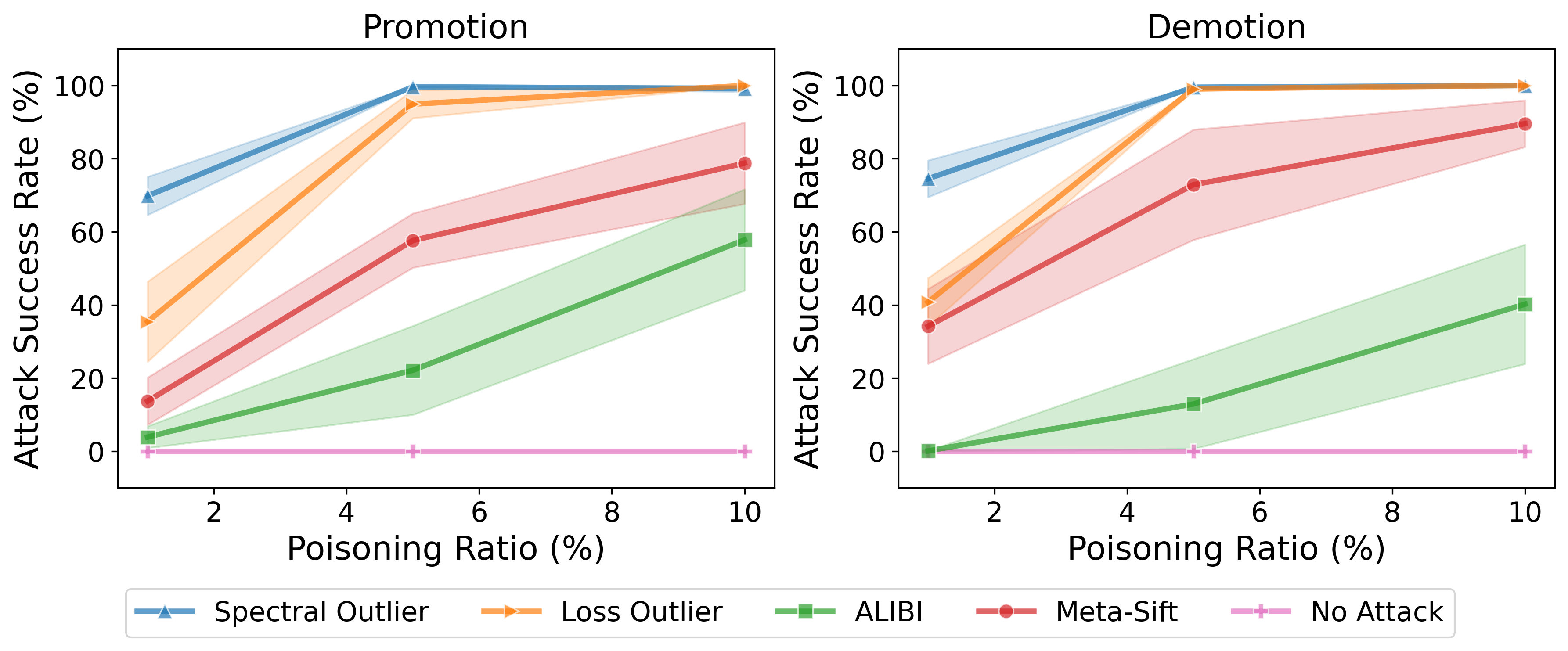}
\caption{Efficacy of defense against promotion (left) and demotion (right) attacks in the Amazon recommendation setting.}
\label{F:amazon_defense}
\end{figure}

\smallskip
\noindent\textbf{Recommendation System.}
In the context of the Amazon recommendation dataset, our results of the relative defense efficacy are shown in Figure~\ref{F:amazon_defense}.
In this setting, we can see the most marked effectiveness of the ALIBI defense, whereas neither spectral outlier nor loss outlier approaches are particularly effective.
For example, even with attack budget at 10\%, the efficacy of the best attack drops from 100\% without defense, to $\sim60$\% after ALIBI defense 
for the promotion attacks, and to $\sim40$\% for the demotion attack.
Nevertheless, this still shows that a significant gap remains in defending against promotion and demotion attacks within our threat model.

%% file: feasibility.tex
\section{Attack Feasibility and Possible Mitigations} 

{Practical exploitation of poisoning attacks remains an open research question. Carlini et al.~\cite{carlini2024poisoning} recently shed light on this by exploiting domain-specific vulnerabilities. In our case of preference poisoning, execution of the attack similarly has to consider how data is collected. To illustrate, the open safety alignment dataset that we use~\cite{ji2024beavertails}, includes $\sim$330K question-answer pairs collected from only 70 annotators. In such cases, even only 4 malicious annotators can poison nearly 6\% of the entire dataset.}

{An important way to mitigate the impact of poisoning attacks explored here is to ensure that datasets are collected from a large number of annotators, with each annotating a relatively small fraction of the data. This would reduce the chances of a small number of malicious labelers significantly subverting reward model training. Similarly, datasets commonly have very few annotators label the same query; we can obtain far greater robustness to poisoning attacks by ensuring that each query has many responses and taking the majority vote.}                      

%% file: conclusion.tex
\section{Conclusion}

The problem of poisoning attacks generally, and label-flipping in particular, has received much attention in the traditional supervised learning settings, such as classification and regression.
Despite some superficial similarity, however, the problem of preference poisoning is structurally distinct, both in terms of the particular optimization problem being solved in learning the reward model, and the particulars of the natural threat model involving promotion and demotion of a target set of outcomes (candidates).
Our analysis yields two high-level takeaways with broad significance to security issues surrounding preference elicitation and value alignment.
The first is the importance of comprehensive vulnerability analysis that involves a diverse set of attack techniques; in particular, we show that which attack among a collection is best can vary greatly by domain and environment---that is, by the dataset involved.
The second is the importance of the particular dataset and data distribution.
Indeed, in insightful recent work, \cite{suya2023distributions} observed that the nature of the data distributions (and, thus, datasets) can be essential to robustness of models to data poisoning.
This issue, which we highlight empirically, clearly deserves considerably more attention.
Moreover, our observation that several state-of-the-art defense techniques fail to provide consistently strong defense in our problem setting suggests that an ability to exert control on the nature of the data we collect
can have substantial consequences for the vulnerabilities we have studied here.

%% file: appendix.tex
\appendices

\section{Additional Experiment Details}
\label{S:additional_exp_details}

\subsubsection*{MuJoCo Control}

We use a feedforward 3-layer neural network with ReLU activation functions, where the mid-layer size is 32. We collect 1250 pairs of trajectories for each environment, and train for $2000$ epochs with learning rate $6.25 \times 10^{-4}$. Training and testing data have the same size. For gradient-based black-box attacks, we use a 3-layer neural network with mid-layer size $16$ for attack.

\subsubsection*{Atari Vision-Based Control} 

We trained for 100 epochs with a $6.25 \times 10^{-4}$ learning rate. 
PCA reduces the input dimension to $20$, and we then concatenate it with the action (final dimension is $21$). Then we use a 3-layer neural network with mid-layer size $16$.

\subsubsection*{Recommendation System}

We explored both a simple linear model and a 3-layer neural network with ReLU activation functions and a mid-layer size of 1024. In the PCA scenario, we reduced it to 20 components and then employed a 3-layer neural network with ReLU activation functions and a mid-layer size of 32. 
Training data has $4500$ pairwise preferences, and testing data has $500$ pairwise preferences. For gradient-based black-box attacks, we first do PCA to reduce its dimension to size $10$; then we use a 3-layer neural network with mid-layer size $16$.

\begin{figure}[htbp]
    \centering
\includegraphics[width=\linewidth]{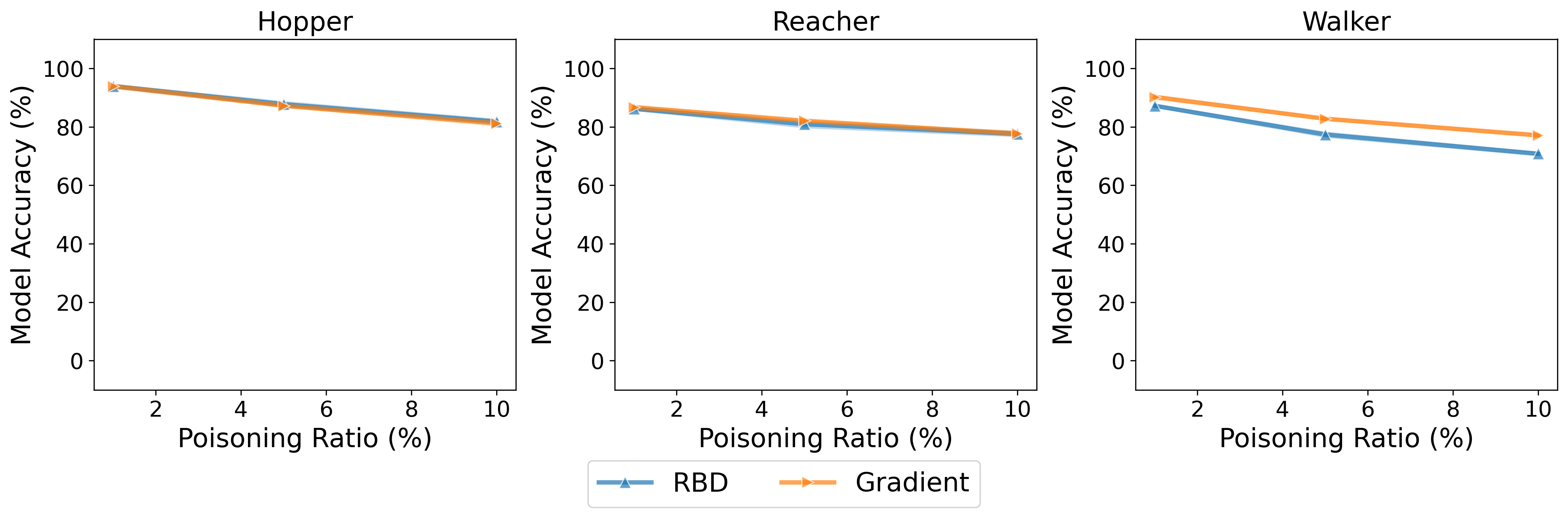}
\caption{Demotion attack stealth in MuJoCo: test set accuracy.}
\label{F:mujoco_demotion_accuracy}
\end{figure}

\begin{figure}[htbp]
    \centering
\includegraphics[width=\linewidth]{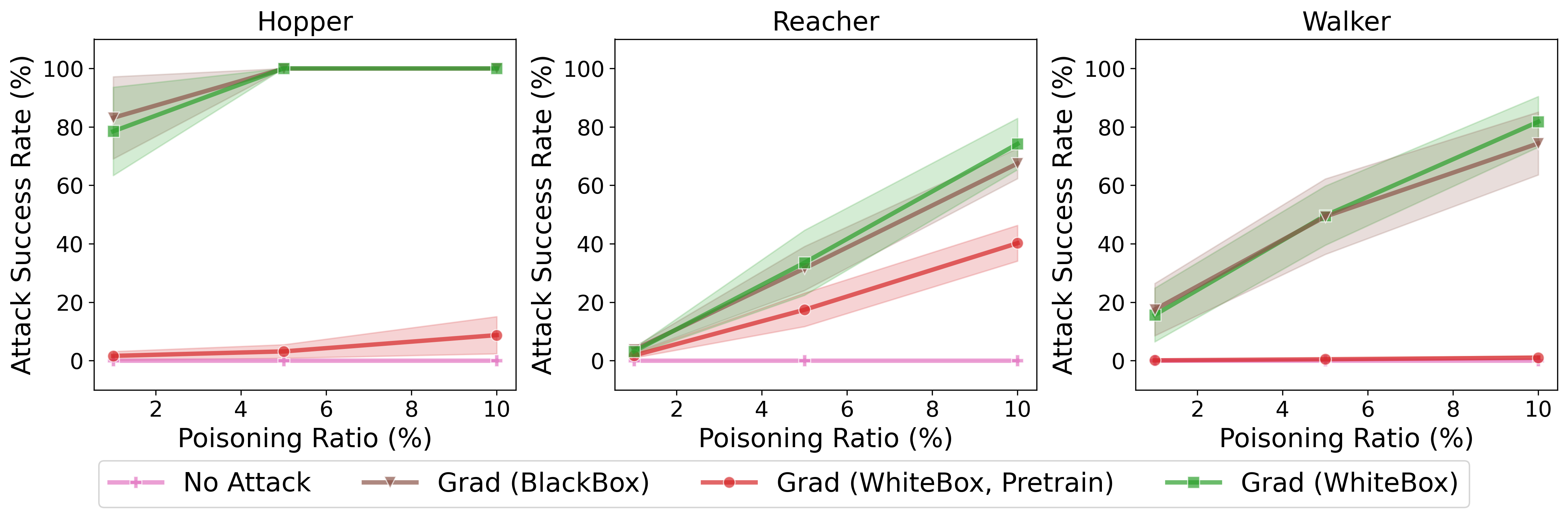}\\
\includegraphics[width=\linewidth]{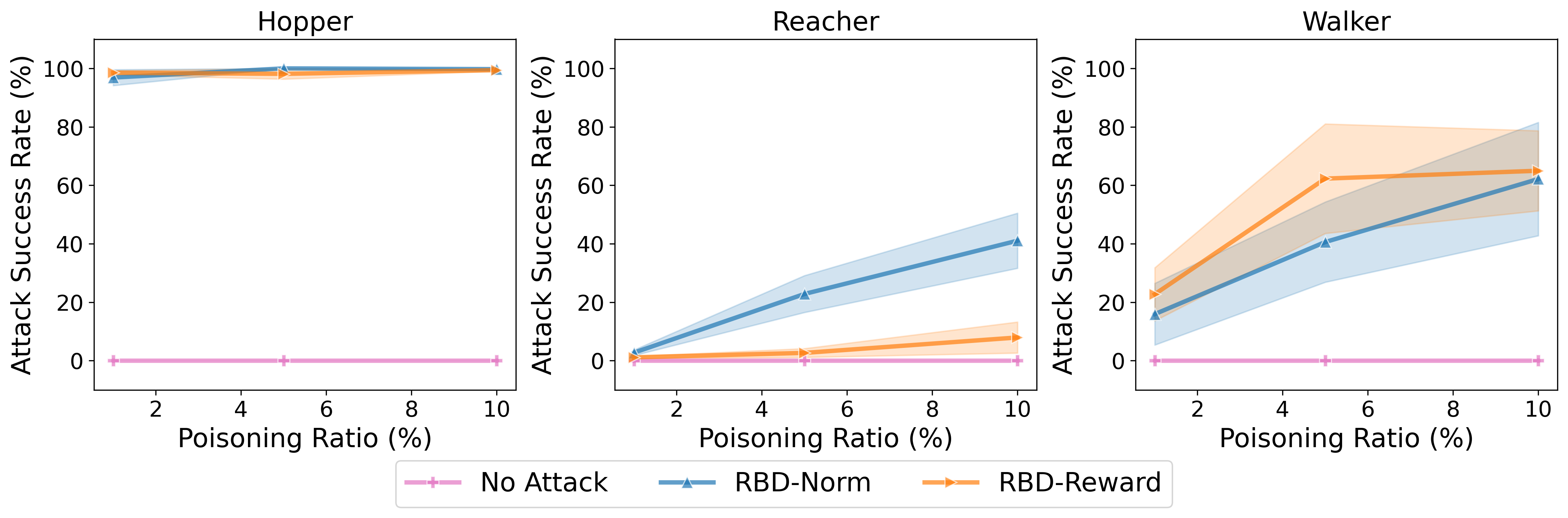}
\caption{Relative efficacy of gradient-based demotion attacks (top row) and RBD demotion attacks (bottom row) in MuJoCo.}
\label{F:mujoco_demotion_ablation}
\end{figure}

\begin{figure}[htbp]
    \centering
    \includegraphics[width=\linewidth]{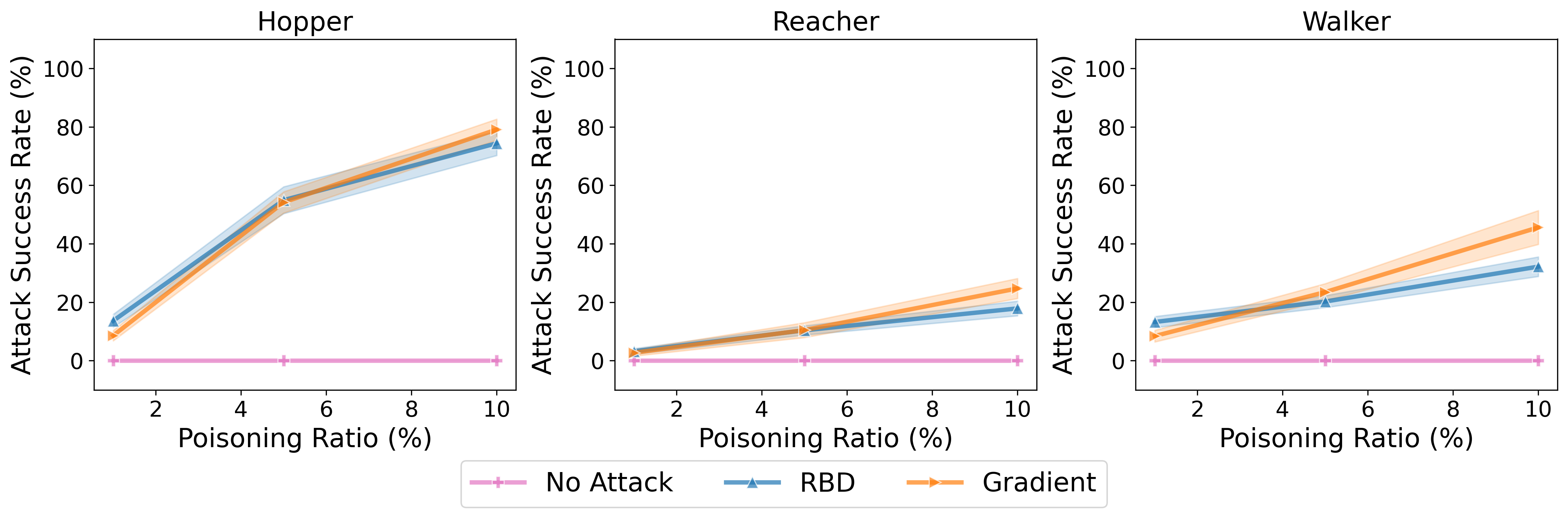}\\
    \includegraphics[width=\linewidth]{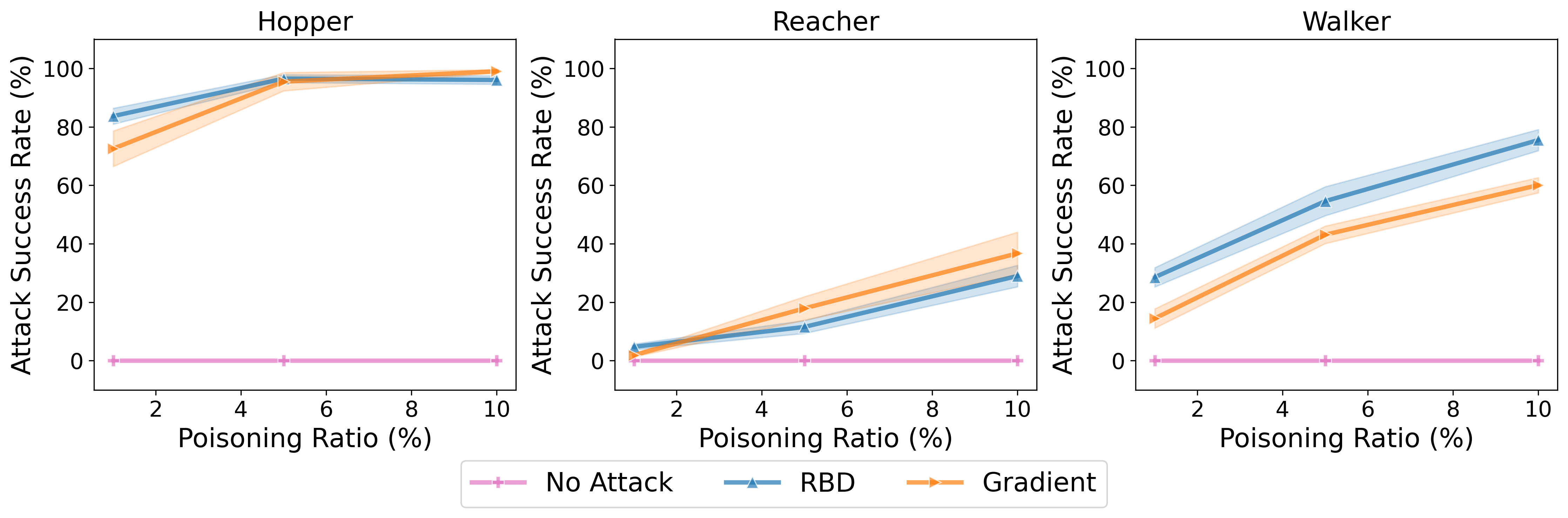}\\
\caption{Attack efficacy in MuJoCo with 5 target candidates.}
\label{F:mujoco_multiple_targets}
\end{figure}

\begin{figure}[htbp]
    \centering
\includegraphics[width=\linewidth]{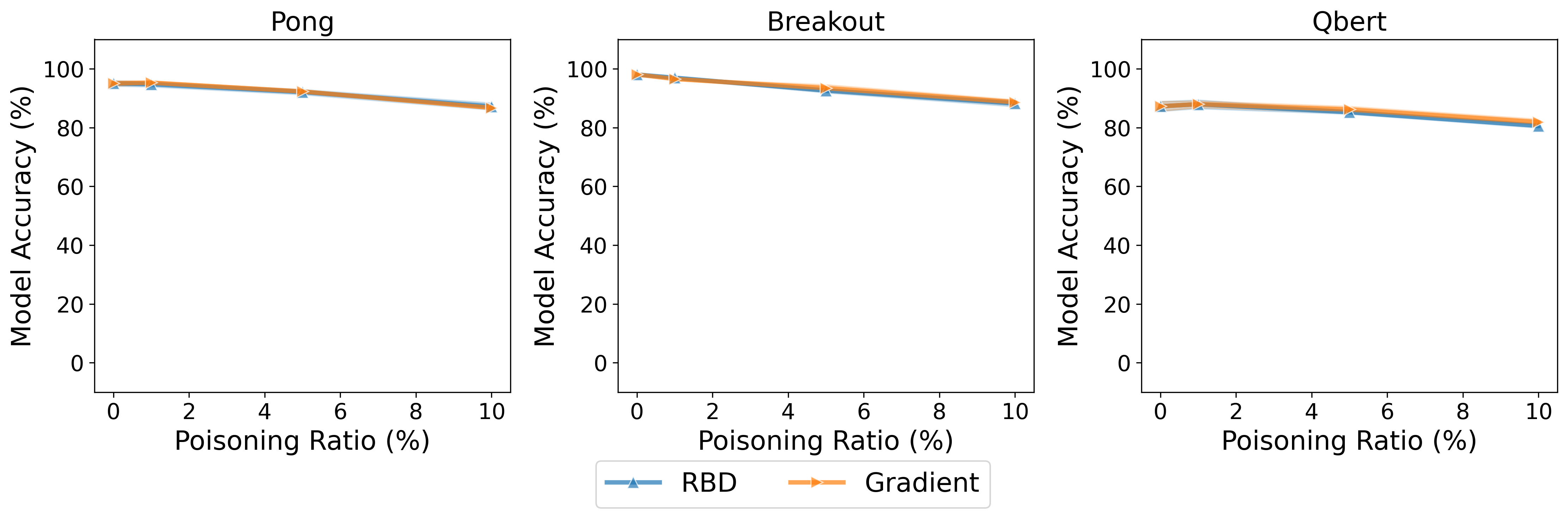}
\caption{Demotion attack stealth in Atari: test set accuracy.}
\label{F:atari_demotion_accuracy}
\end{figure}

\begin{figure}[!t]
    \centering
\includegraphics[width=\linewidth]{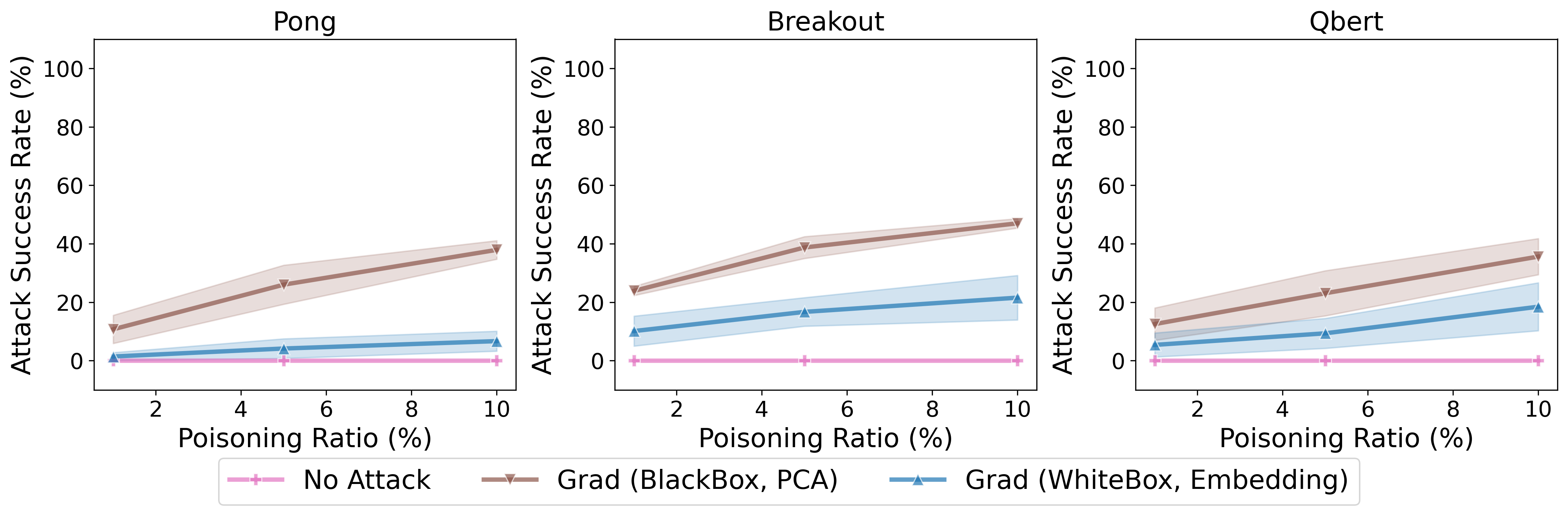}\\
\includegraphics[width=\linewidth]{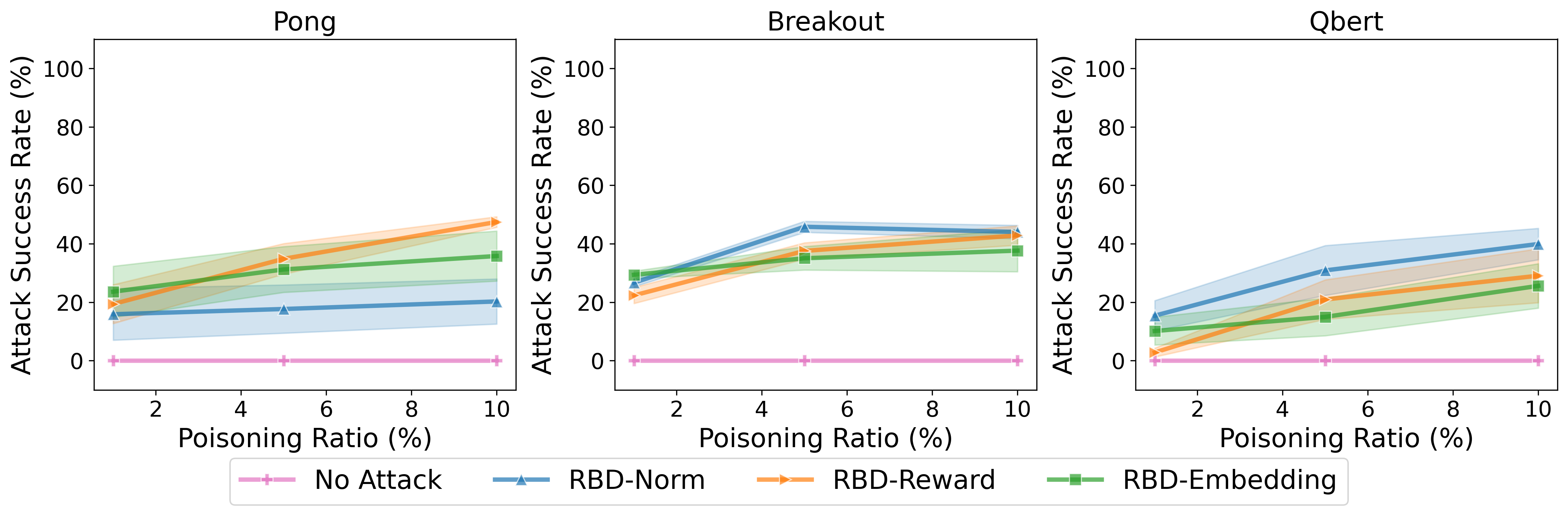}
\caption{Relative efficacy of gradient-based demotion attacks (top row) and RBD demotion attacks (bottom row) in Atari.}
\label{F:atari_demotion_ablation}
\end{figure}

\begin{figure}[htbp]
    \centering
    \includegraphics[width=\linewidth]{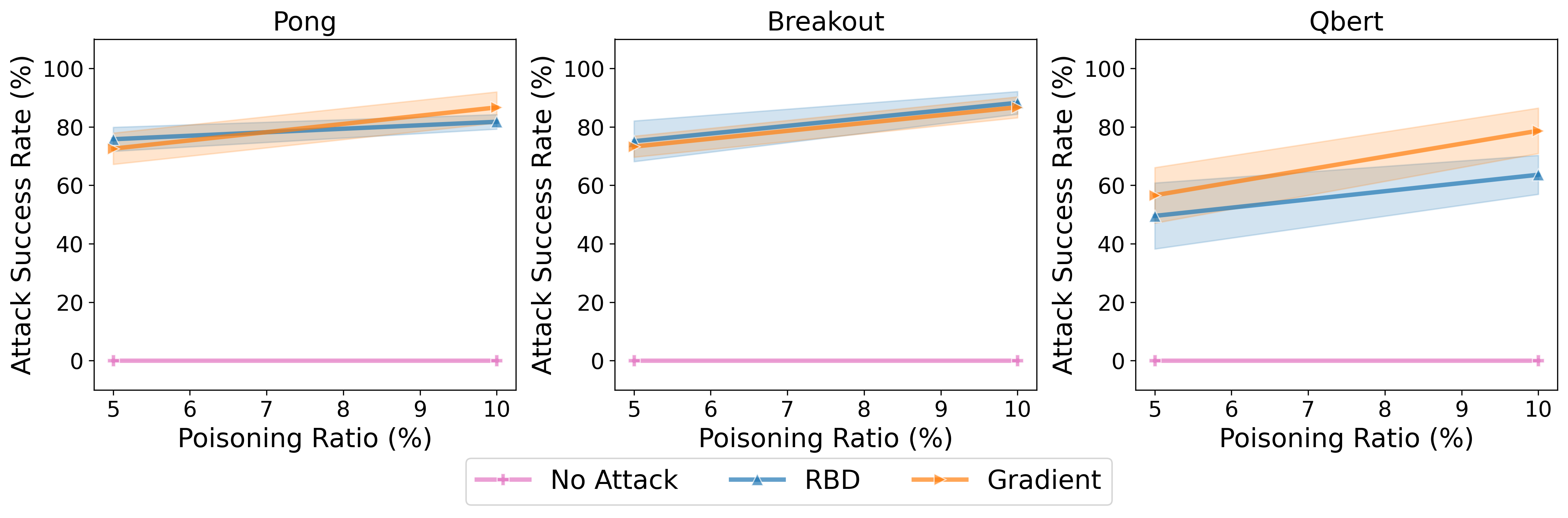}\\
    \includegraphics[width=\linewidth]{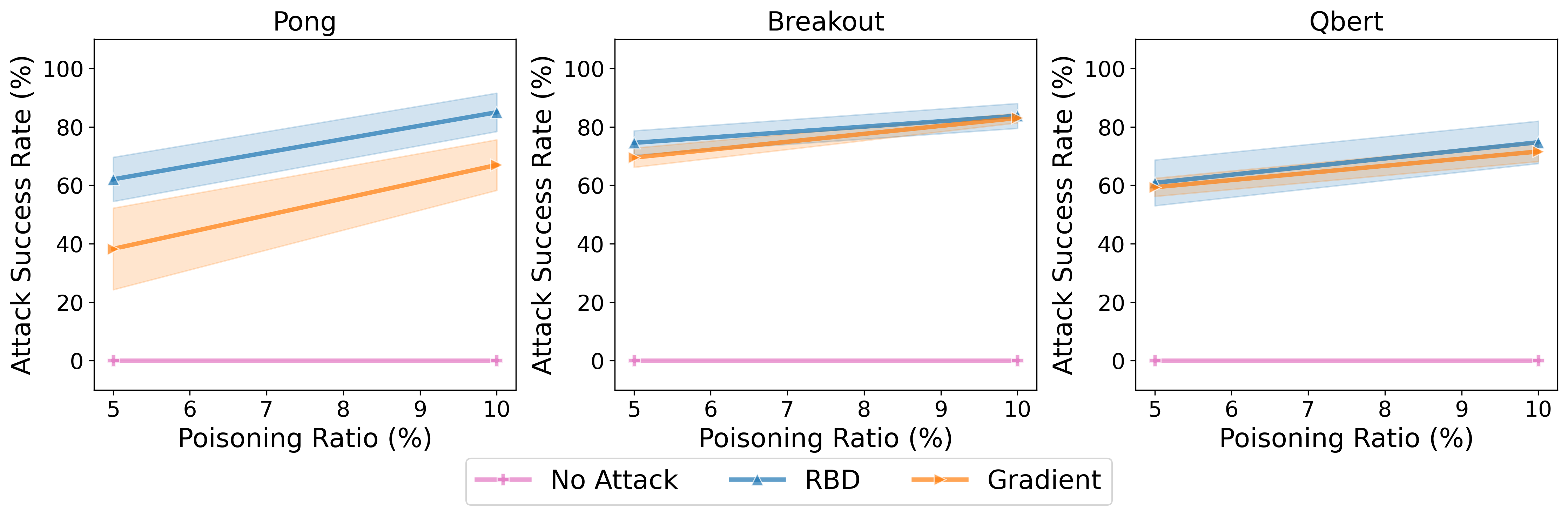}\\
\caption{Attack efficacy in Atari with 5 target candidates.}
\label{F:atari_multiple_targets}
\end{figure}

\begin{figure}[htbp]
    \centering
    \includegraphics[width=0.9\linewidth]{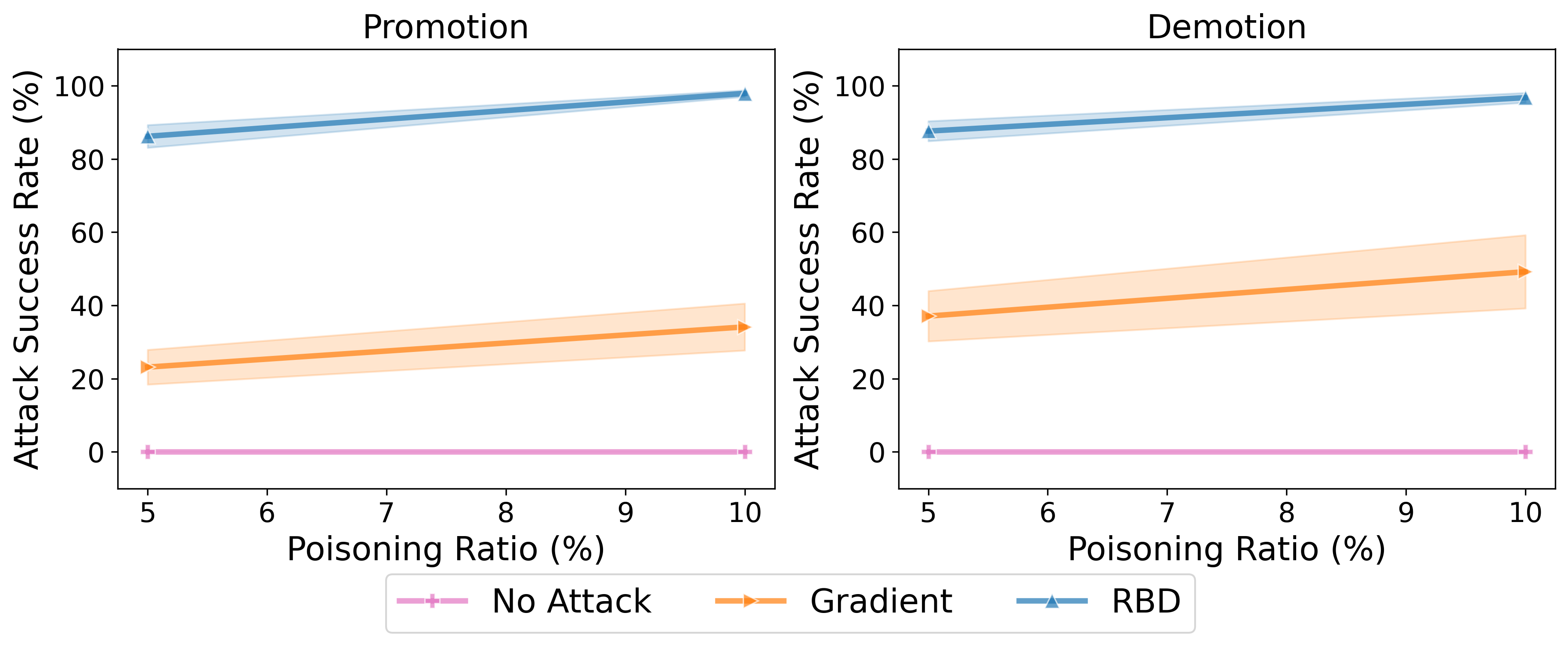}\\
    \includegraphics[width=0.9\linewidth]{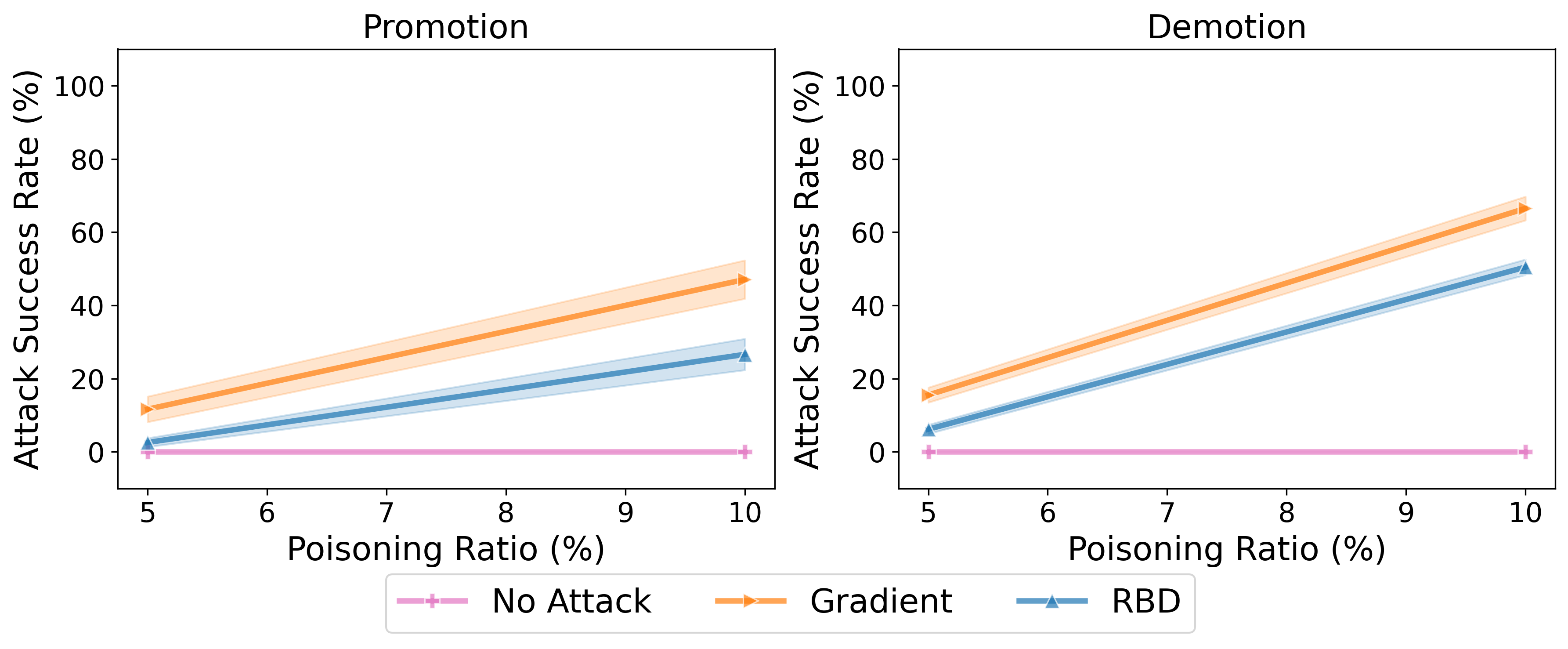}
\caption{Attack efficacy with 5 target candidates on the Amazon recommendation ratings dataset. Neural Network Model (top row) and Linear Model (bottom row).}
\label{F:amazon_multiple_targets}
\end{figure}

\section{Additional Results}

\subsubsection*{MuJoCo Control} 
Figure~\ref{F:mujoco_demotion_accuracy} presents the results of trained reward model accuracy for demotion attacks.
The results are similar to promotion attacks: accuracy degradation is small, demonstrating that the attack is sufficiently stealthy.
Figure~\ref{F:mujoco_demotion_ablation} presents the ablation results for gradient and RBD methods in the case of demotion attacks.
Finally, Figure~\ref{F:mujoco_multiple_targets} shows that the results for multiple target candidates (5, in this case) are comparable to a single target candidate.

\subsubsection*{Atari Vision-Based Control} 
Figure~\ref{F:atari_demotion_accuracy} presents the results of trained reward model accuracy for demotion attacks in Atari.
We can again observe that there is little degradation in training accuracy after the attack.
Figure~\ref{F:atari_demotion_ablation} presents the ablation results for gradient and RBD methods in the case of demotion attacks in the Atari domain.
Figure~\ref{F:atari_multiple_targets} shows that the results for multiple target candidates are comparable to a single target candidate in the Atari domain.

\subsubsection*{Recommendation System} 
Figure~\ref{F:amazon_multiple_targets} presents the results with 5 target candidates on the Amazon dataset.
As we can see, these are broadly consistent with the single-target case in the main body.